\documentclass[acmsmall,screen,nonacm]{acmart}
\usepackage{microtype}
\usepackage{graphicx}
\usepackage{tikz-uml}
\usepackage{booktabs} 
\usepackage{natbib}

\usepackage{makecell}
\usepackage{multirow}
\usepackage{lscape}
\usepackage{longtable}
\usepackage{adjustbox}
\usepackage{bbold}
\usepackage{lipsum}
\usepackage{tabularx}
\newcolumntype{F}{>{\RaggedRight\arraybackslash\hsize=1.2\hsize}X}
\newcolumntype{L}{>{\bfseries\RaggedLeft\arraybackslash\hsize=0.8\hsize}X}

\usepackage[normalem]{ulem}

\usepackage{amsmath}
\usepackage{pdflscape}
\usepackage{mathtools}
\usepackage{amsthm}
\usepackage{tikz}
\usetikzlibrary{arrows.meta, positioning}
\usepackage{subcaption}

\usepackage{enumitem}

\theoremstyle{plain}

\theoremstyle{definition}

\theoremstyle{remark}

\usepackage[textsize=tiny]{todonotes}

\usepackage{colortbl}
\definecolor{LightGray}{gray}{0.9}
\newcommand{\graycell}{\cellcolor{LightGray}}

\title{A survey on Clustered Federated Learning: Taxonomy, Analysis and Applications}
\titlenote{This manuscript is a major revision of the version previously deposited as https://arxiv.org/abs/2501.17512 arXiv:2501.17512. Updates include a restructuring of the taxonomy, a literature review update through June 2025, and an adjustment to the authorship order to reflect contributions to this revised version.}
\author{Michael Ben Ali}
\email{Michael-Eddy.Ben-Ali@irit.fr}
\affiliation{%
  \institution{Université Toulouse, UT3, IRIT, CNRS}
  \city{Toulouse}
 \country{France}
}
\author{Omar El-Rifai}
\email{Omar.El-Rifai@irit.fr}
\affiliation{%
  \institution{Université Toulouse, UT3, IRIT, CNRS}
  \city{Toulouse}
  \country{France}
}
\author{Imen Megdiche}
\email{imen.megdiche@irit.fr}
\affiliation{%
  \institution{INU Champollion, ISIS Castres, IRIT, CNRS}
  \city{Castres}
  \country{France}
}
\author{André Peninou}
\email{andre.peninou@irit.fr}
\affiliation{%
  \institution{Université Toulouse, UT2J, IRIT, CNRS}
  \city{Toulouse}
  \country{France}
}
\author{Olivier Teste}
\email{olivier.teste@irit.fr}
\affiliation{%
  \institution{Université Toulouse, UT2J, IRIT, CNRS}
  \city{Toulouse}
  \country{France}
}
\begin{document}

\begin{abstract}
As Federated Learning (FL) expands, the challenge of non-independent and identically distributed (non-IID) data becomes critical. Clustered Federated Learning (CFL) addresses this by training multiple specialized models, each representing a group of clients with similar data distributions. However, the term "CFL" has increasingly been applied to operational strategies unrelated to data heterogeneity, creating significant ambiguity. This survey provides a systematic review of the CFL literature and introduces a principled taxonomy that classifies algorithms into Server-side, Client-side, and Metadata-based approaches. Our analysis reveals a distinct dichotomy: while theoretical research prioritizes privacy-preserving Server/Client-side methods, applied papers overwhelmingly favor Metadata-based approaches, prioritizing efficiency over privacy. Furthermore, we explicitly distinguish "Core CFL" (grouping clients for non-IID data) from "Clustered X FL" (operational variants for system heterogeneity). Finally, we outline lessons learned and future directions to bridge the gap between theoretical privacy and practical efficiency.
\end{abstract}
 
\maketitle

\section{Introduction}

\textit{Federated Learning (FL)} was designed to collaboratively train a \textbf{single global} deep learning model — that is, one shared set of parameters collaboratively optimized by all participating clients — without sharing raw data across devices. In a typical FL setup, a central server distributes a model to participating clients, who train it locally using private data. The clients return their updated parameters\footnote{For convenience, we refer to the model and its parameters interchangeably. We assume the model architecture is fixed; thus, references to the parameters define the model.} to the server, which aggregates them into a new global model. This complete cycle of distribution, local training, and aggregation constitutes a single \textbf{communication round}. The process iterates over multiple rounds until convergence. FL is designed for large-scale, distributed environments where communication is constrained, and data is decentralized, heterogeneous, and privacy-sensitive \citep{mcmahan2017communication}.

Nevertheless, subsequent studies have revealed several shortcomings in the initial approaches \citep{kairouz2021advances}. FL solutions are often inadequate when faced with data heterogeneity in many practical applications \cite{ye2023heterogeneous,pfeiffer2023federated}. Consequently, particular attention is given to the challenges of training a single global solution when datasets are heterogeneous (non-IID) across clients. 

Zhao et al. \citet{zhao2018federated} demonstrated the difficulty of this scenario by showing 
that the optimization direction of federated models on non-IID data diverges significantly from that of a hypothetical centralized model trained on the same data. 
This weight divergence creates a larger gap between the federated and optimal centralized weights than in homogeneous settings, resulting in a significant 
reduction in performance.

In response to this challenge, multiple variants of FL have emerged in the scientific literature \citep{neurocomputingsurvey,fxl}, including model aggregation techniques, knowledge distillation, regularization-based methods, and personalized FL. One such approach is known as \textit{Clustered Federated Learning (CFL)} \citep{IFCA, sattler2020clustered}, which relaxes the constraint of a single global model by training multiple models, each tailored to a group of clients 
with similar data distributions. This strategy builds upon the analysis of weight divergence in Zhao et al.~\citep{zhao2018federated}, leveraging the observation that clients with similar local data distributions tend to update their models in similar directions. The CFL approach assumes the existence of numerous, 
distinct distributions of client data. It aims to capture this structure by grouping clients and learning \textit{an independent model for each cluster}.

\textbf{The goal of the paper is to conduct a Survey of Clustered Federated Learning (CFL)}. From our literature review, however, it is clear that the term CFL has not remained confined to its original meaning, that is, to address non-IID data by training multiple models. Yet, over time, other uses of the same terminology appeared in the literature. For example, several studies employ the term CFL to describe solutions using 
clustering to tackle system heterogeneity 
\cite{feng2022mobility,zhou2023hierarchical,wang2024social,wen2024dynamic,solat2023novel,yang2024clustering}. 
However, these works retain the single global model objective rather than 
addressing non-IID data distributions through multiple models. Furthermore, 
significant terminological inconsistency exists in the literature: terms such 
as \textit{Clustered FL}, \textit{Clustering FL}, \textit{Cluster-based FL}, and \textit{Clustering-based FL} are often used interchangeably, and sometimes overlap with unrelated notions such as Federated Clustering. 

\textbf{One of the objectives of this survey is therefore to disentangle the different meanings and research trends} of CFL in the literature. We will explicitly distinguish what we call core CFL: algorithms that cluster clients to tackle non-IID data by training multiple models. We also separate other uses we call \textbf{Clustered X FL}, where clustering supports goals such as communication efficiency, client scheduling, or system-level optimization without addressing non-IID data. This distinction structures our survey and ensures clarity in positioning CFL relative to other branches of the FL literature.

\paragraph{\textbf{Organization of the Content of the Paper.}} 
The survey is structured to provide a unified framework for understanding CFL and its applications. 
First, Section ~\ref{sec:methodology} presents our literature search methodology. Section~\ref{sec:positionning} then discusses the positioning of CFL within the broader field of Federated Learning.
Section~\ref{sec:core} introduces Core CFL, beginning with background on FL and the non-IID challenge (Subsection~\ref{sec:fl}). Subsection~\ref{sec:cfltaxonomy} formalizes the CFL problem and presents our core taxonomy, classifying state-of-the-art domain agnostic algorithms into \textbf{Server-side}, \textbf{Client-side}, and \textbf{Metadata-based} approaches. This is supported by a classification table in Section~\ref{sec:clf_classification} summarizing existing methods. We then highlight common design challenges, such as determining the number of clusters, scalability, soft clustering, dynamic environments, security, and privacy (section~\ref{sec:number_of_clusters} to~\ref{sec:privacy-meta}). Section~\ref{sec:workflow} introduces a unified multi-tier workflow that abstracts the structure of CFL algorithms. Section~\ref{sec:cfl_eval} discusses CFL evaluation methodologies, covering the role of the non-IID data taxonomy in creating benchmark configurations. Section~\ref{sec:app} reviews the application studies of Core CFL across domains like IoT, Mobility, Energy, and Healthcare. Section~\ref{sec:other_use_cases} distinguishes Core CFL from \textbf{Clustered X FL} variants, which apply clustering for operational optimization, often maintaining a single global model. Section~\ref{sec:future_directions} identifies open challenges and future 
research directions. Finally, Section~\ref{sec:conclusion} concludes the survey by summarizing key findings and highlighting the distinction between 
CFL uses.

 \section{Systematic Literature Search Methodology}\label{sec:methodology}
To ensure a rigorous review of the CFL literature, we adopted a systematic methodology for paper selection, guided by clear inclusion criteria. The scope of the review was limited to publications released before June 2025. We constructed a search query targeting CFL terminology :
\begin{quote}
\footnotesize
\texttt{("Clustered" OR "Cluster-based" OR "Clustering-based" OR "Clustering" OR } \\ 
\texttt{ Clustered FL" OR "Cluster-based FL" OR "Clustering-based FL" OR "Clustering FL")}\\ 
\texttt{AND "Federated Learning" } 
\end{quote} 

This query was applied across four major databases: 
Web of Science, Semantic Scholar, Google Scholar, and Dimensions.ai. These platforms were chosen both for their free accessibility and coverage of computer science and interdisciplinary research on FL.

Only peer-reviewed articles published in journals or conferences indexed in Scopus or Web of Science were retained. Workshop papers and non-peer-reviewed sources, such as preprints, were discarded. To ensure quality, ranking filters were applied according to the year of publication: journal articles were required to appear in Q1-ranked journals based on the Scimago Journal Rank (SJR), while conference papers were selected only if classified as A or A* in the CORE Conference Ranking. For core CFL papers introducing new domain-agnostic algorithms (see Section~\ref{sec:clf_classification}), all selected works provide experimental evaluations against prior CFL approaches to ensure comparability and relevance. Exceptions were made for foundational papers \citep{ghosh2019robust,sattler2020clustered,mansour2020three,
briggs2020federated,IFCA} which established the CFL paradigm. In total, 529 articles were initially retrieved from queries. After removing duplicates, extended versions, false positives, and papers out of scope, 130 CFL articles were retained. Among these, 36 correspond to \textit{domain agnostic core CFL} (Q1/A/A*) publications, while 75 are \textit{application-oriented} and 19 classified as \textit{Clustered X FL}.

\section{Positioning of CFL within the Federated Learning Landscape}
\label{sec:positionning}
Non-IID data remains a major challenge in FL \citep{kairouz2021advances,fxl,neurocomputingsurvey}. Several strategies have been proposed to mitigate heterogeneous client distributions. These include techniques that modify the optimization dynamics, impose regularization or model distillation (All highlighted in Ji et al. \cite{fxl}). These approaches share the single global model assumption. When data distributions are only mildly heterogeneous, such strategies often suffice to stabilize training and preserve convergence. However, when client distributions are deeply distinct, the single-model assumption becomes restrictive. Updating one global model to fit multiple divergent updates leads to unstable optimization and poor performance.

\begin{figure}[htbp!]
    \centering
    \begin{subfigure}[t]{0.48\textwidth} 
        \centering

        \includegraphics[height=5.5cm, keepaspectratio]{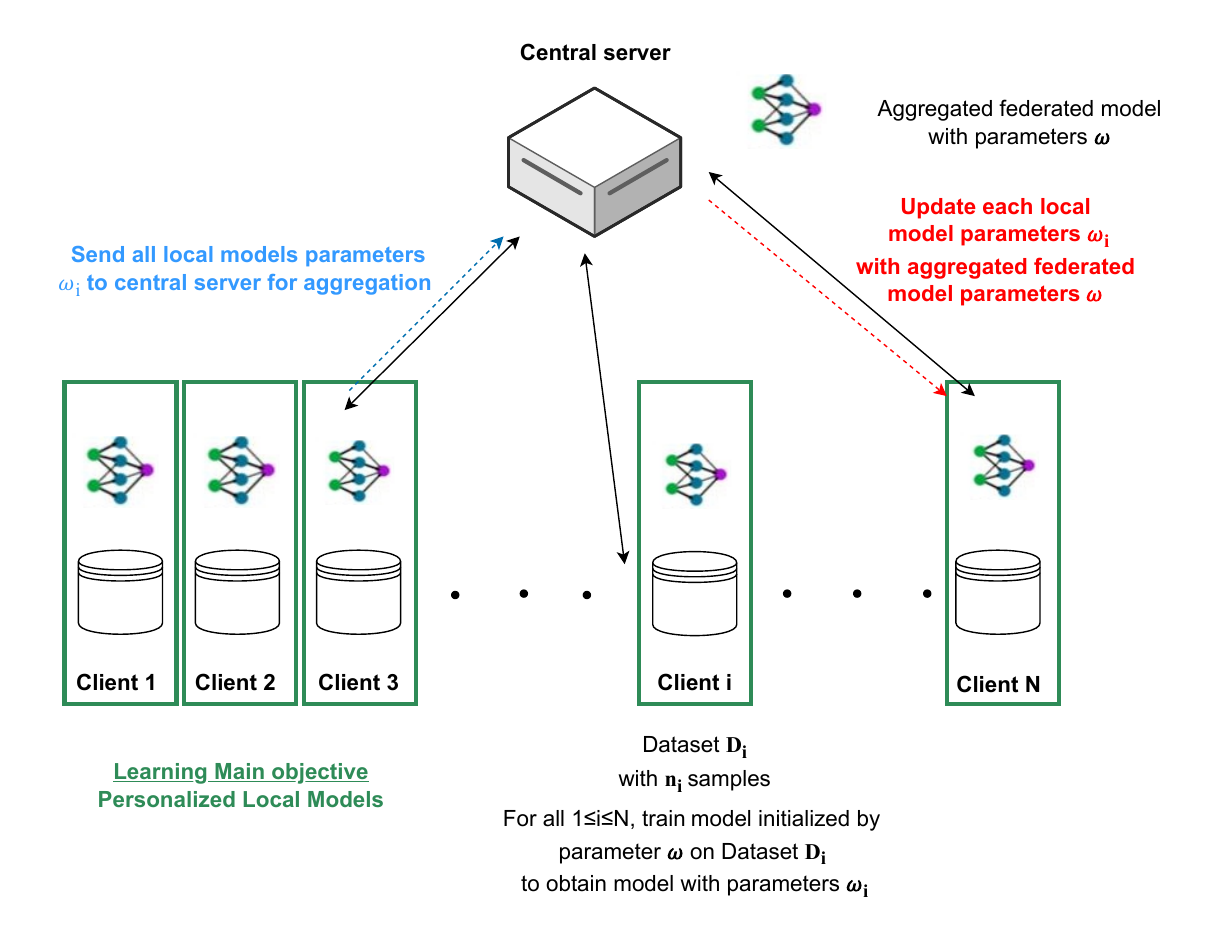}
        \caption{PFL: Each client trains a personalized model $\omega_i$ starting from shared parameters model $\omega$. Unlike standard FL, the learning objective shift to local.}
        \label{fig:PFL}
    \end{subfigure}
    \hfill 
    \begin{subfigure}[t]{0.48\textwidth}
        \centering

        \includegraphics[height=4.7cm, keepaspectratio]{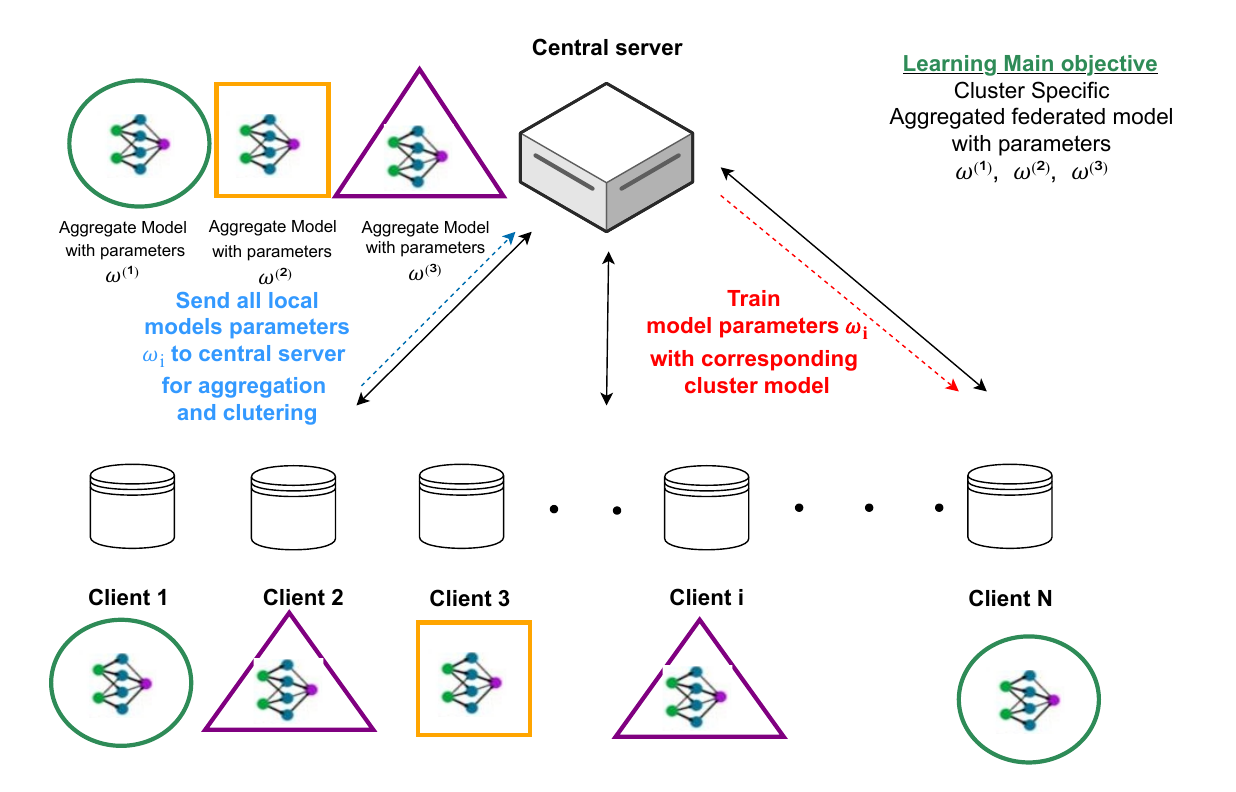}
        \caption{CFL: Clients are grouped into clusters (triangle, circle, square). Each cluster trains separate parameters $\omega^{(k)}$, rather than a single global model.}
        \label{fig:CFL}
    \end{subfigure}
    
    \caption{Comparison between Personalized Federated Learning (PFL) and Clustered Federated Learning (CFL).}
    \label{fig:pfl_cfl_comparison}
\end{figure}

CFL addresses this limitation by assuming the existence of multiple client groups, each following its own underlying distribution. The goal is not to correct clients so that they all fit a shared model but instead to identify coherent subsets of clients and train one model per group. CFL therefore replaces the single-model constraint with multiple models adapted to different distributions.
While few categorize CFL as a subfield of Personalized Federated Learning (PFL) (a family of techniques that relaxes the single global model constraint by tailoring models to individual clients \cite{personalizesurvey, PFLsurvey}), most research treats CFL as a distinct approach \citep{neurocomputingsurvey,fxl}. The confusion often arises from a personalization standpoint. In PFL, the personalization goal is typically associated with optimizing $N$ models locally for $N$ clients (Figure~\ref{fig:PFL}) using one single global model. In CFL, by contrast, the objective is to optimize $N$ models using only $K$ unique models ($K \ll N$), since clients within the same cluster share a common model (Figure~\ref{fig:CFL}). We suggest that CFL is better viewed as distinct from PFL, as the two approaches rely on different assumptions about the underlying data distribution across clients. While PFL typically assumes a single global distribution and adapts the model to individual clients to accommodate local empirical distribution differences, CFL assumes the existence of multiple distinct client groups, each characterized by its own distribution. This perspective aligns with how some popular FL surveys \citep{neurocomputingsurvey,fxl} present the field, where CFL and PFL are introduced as separate approaches. Moreover, some CFL algorithms \citep{FedCE, FedCAM} integrate personalization techniques to tackle intra-cluster heterogeneity, optimizing both $K$ global models and $N$ local models. This further highlights the need to treat CFL and PFL as separate lines of research. Since the focus of this survey is CFL, we suggest readers look at Sabah et al.\cite{PFLsurvey}, Liu et al. \cite{neurocomputingsurvey} and Ji et al. \cite{fxl} for more details on PFL. 

Finally, it is also important not to confuse CFL with what is usually called Federated Clustering  \cite{fxl} (sometimes called unsupervised federated learning). While both involve clustering techniques, Federated Clustering usually refers to distributed clustering methods applied directly to unlabeled data across clients, where the task is unsupervised. On the other hand, the objective of the CFL is not a clustering task in itself; clustering is only used as a mechanism to group clients' models for more effective training. The CFL is designed as agnostic to the learning tasks. Its function does not modify the FL protocole making it compatible with all task compatible with standard FL.

\section{Core Clustered Federated Learning for Non-IID Data} \label{sec:core}

To establish a clear frame of reference for further discussions, we begin by outlining the FL framework described in \citep{mcmahan2017communication} and by specifying the Non-IID data problem as mentioned in that study. Before proceeding, it is important to clarify our article selection criteria for this section. We focus here exclusively on what we call \textbf{core CFL}—that is, algorithms designed to handle non-IID data.

\begin{table}[htbp]
    \centering
    \scriptsize
    \begin{tabular}{c p{0.9\textwidth}}
        \hline
        \textbf{Notations} & \textbf{Description} \\
        \hline
        $N$ & Number of clients in the FL ecosystem, $N\geq 2$ \\
        $I$ & $I = \{1, \ldots, N\}$ set of all clients\\ 
        $i$ & Index of client ($i \in I$) \\      
        $D_i$ & Local dataset of client $i$ noted as $D_i = \left\{(x^{(i)}_j, y^{(i)}_j) \mid 1 \leq j \leq n_i\right\}$, where $(x^{(i)}_{j})_{1 \leq j \leq n_i}$ and $(y^{(i)}_{j})_{1 \leq j \leq n_i}$ represents respectively features and target \\
        $n_i$ & Number of samples in $D_i$. We note $|D_i| = n_i$ \\
        $d$ & Number of parameters of considered neural networks (NN), $d \in \mathbb{N}^*$\\ 
        $\omega$ & Federated neural network model parameters with $\omega \in \mathbb{R}^d$ \\
        $\omega_i$ & Local neural network parameters of client $i$ with $\omega_i \in \mathbb{R}^d$ \\
        $f$ & FL objective function \\
        $f_i$ & Local objective function for client $i$ defined to optimize a local model on $D_i$\\
        $(\Omega, \mathcal{F}, \mathcal{P})$ & Probability space where  $\Omega$ is the sample space, representing all possible outcomes of the random experiment associated with sampling from the dataset $D= \cup_{i=1}^{N}D_i$ \\
        &$\mathcal{F}$ the associate sigma algebra and $\mathcal{P}$ the probability measure \\ 
        $X$ & Random variable of the possible outcome of features of $D$. In this case, each $(x^{(i)}_{j})_{1 \leq j \leq n_i}$ for $i \in I$ is a realization of $X$. \\
        $Y$ & Random variable of the possible outcome of targets of $D$. In this case, each $(y^{(i)}_{j})_{1 \leq j \leq n_i}$ for $i \in I$ is a realization of $Y$. \\
        $(X,Y)$ & Joint random variable of the possible outcome of features and targets of $D$. We have $(X,Y) : \Omega \mapsto D$ \\
        $\mathcal{P}_{i}(X)$ & Marginal distribution of dataset $D_i$ features \\
        $\mathcal{P}_{i}(Y)$ & Marginal distribution of dataset $D_i$ target \\
        $\mathcal{P}_{i}(X, Y)$ & Joint distribution of features and targets of dataset $D_i$ \\
        $\mathcal{P}_{i}(X | Y)$ & Conditional distribution of features conditioned by target of dataset $D_i$ \\
        $\mathcal{P}_{i}(Y | X)$ & Conditional distribution of targets conditioned by features of dataset $D_i$ \\
        $K$ & Number of clusters, $K \leq N$\\
        $C_k$ & $k$-th cluster with $k \in \{1, \ldots, K\}$; a cluster $C_k$ will be considered as a subset of $I$. Meaning that $C_k \subseteq I $ and $I = \bigcup\limits_{k=1}^{K}C_k$ \\
        $F_k$ & Objective function to optimize the federated model on cluster $C_k$ \\
        $\mathrm{dist}(\cdot \hspace{1pt}, \cdot)$ & Generic distance metric \\
        $\mathbb{1}_{i \in C_k}$ & Equal to 1 if client $i$ is in cluster $C_k$, 0 otherwise \\
        $\mu_k$ & Cluster $C_k$ representative point essential for cluster formation; a vector of size $d$ calculated using weights $\omega_i$ for all $i \in C_k$ \\
        $\eta_i$ & Metadata parameters of client $i$; may correspond to local data statistics, distribution statistics, or any type of metadata representation of client $i$ (see Section \ref{sec:metadata-based}) \\
        $\delta_k$ & Cluster $C_k$ metadata-based representative point essential for cluster formation; This point share the same type as metadata parameters $\eta_i$ for all clients $i \in C_k$ (see Section \ref{sec:metadata-based}) \\
        \hline
    \end{tabular}
    \caption{Summary of notations}
    \label{tab:notations}
\end{table}

\subsection{Background on Federated Learning and Data Heterogeneity} \label{sec:fl}

\textit{Federated Learning (FL)} is an iterative method to train a model on a network of devices without sharing raw data. We consider a distributed setting with $N$ clients\footnote{In this study, we use terms clients or devices interchangeably when speaking about FL participants.}, $N \geq 2$, coordinated by a central server. Each client $i \in I = \{1,\ldots,N\}$ holds a local dataset $D_i$ that cannot be shared with others. The goal of federated learning is to find a set of parameters $\omega \in \mathbb{R}^d$, where $d$ is the number of model parameters, that minimizes a global objective function $f(\omega)$ based on all local datasets, all this without sharing $D_i$, $\forall i \in I$ to ensure data privacy. Letting $n_i = |D_i|$ denote the dataset size of client $i$ and $n = \sum_{j=1}^{N} n_j$ the total number of samples across all clients. Formally, this global objective can be expressed as:
\begin{equation}\label{eq:fl_global_objective}
\min_{\omega \in \mathbb{R}^d} f(\omega) := \sum_{i=1}^N \frac{n_i}{n} f_i(\omega),
\end{equation}
where $f_i(\omega)$ denotes the local objective function of client $i$. In supervised learning, one typically defines
\begin{equation}\label{eq:fiomega}
f_i(\omega) = \mathbb{E}_{(X,Y) \sim D_i}[L_{i}(X,Y,\omega)],
\end{equation}

that is, the expected loss $L_i$ over samples $(X,Y)$ drawn from the client’s data distribution $D_i$. This formulation is general and applies to a wide range of machine learning models. To solve this objective without compromising privacy, \citet{mcmahan2017communication} introduced the FedAvg algorithm. In this protocol, during multiple iterative rounds, each client optimizes $f_i(\omega)$ locally to find parameters $\omega_i$, and the server aggregates the resulting models through the weighted averaging defined in Equation (\ref{eq:fl_global_objective}). The FedAvg protocol propose that only $\omega_i$ (local model parameters) and $n_i$ (local dataset size) can be shared by each client with the server to maintain privacy.

A FL network consists of multiple distributed clients, each owning data that reflects its local context (usage patterns or environment). As a result, the statistical properties of those datasets are inherently heterogeneous: the distributions of $D_i$ across clients may differ significantly. This heterogeneity, often referred to as non-IID data, is a central challenge for FL~\cite{kairouz2021advances}. It arises naturally in massively distributed environments where each client captures only a narrow and biased view of the overall population distribution. This statistical heterogeneity has direct implications for the optimization process in FL. When clients train on data drawn from different distributions, their local updates $\omega_i$ may point in conflicting directions, making the aggregation of models into a single global parameter vector $\omega$ suboptimal, affecting performance. To tackle the non-IID problem, CFL proposes grouping clients with similar distributions into clusters and training a shared model for each cluster.

\subsection{Clustered Federated Learning Formulation and taxonomy}\label{sec:cfltaxonomy}
We define a generic CFL optimization formulation. Let $N$ clients $i \in \{1,\ldots,N\}$ having each a dataset $D_i$ containing $|D_i|= n_i$ data samples. We assume that those samples are derived from  $K$ different data probabilistic distributions $\mathcal{P}_1,\ldots, \mathcal{P}_K$ with $1 \leq  K < N$.
This implies that the clients can be partitioned into $K$ disjoint clusters $C_1,\ldots, C_K$ where the clients' dataset distributions are homogeneous intra-cluster. For the sake of simplicity in notations, each cluster will be considered as a partition of $I =\{1,\ldots,N\}$ with the notation $i\in C_k$ meaning that the client of index $i$ is inside cluster $C_k$ (i.e.: $I = \bigcup\limits_{k=1}^{K}C_k$). We would then have $K$ different clusters and for $k \in \{1,\ldots,K\}$ a corresponding objective function $F_k$ to optimize the FL model on cluster $C_k$ of the form:
\begin{equation}
\min_{\omega\in\mathbb{R}^d} F_{k}(\omega) := \sum_{i\in C_k}\frac{n_i}{\sum_{j\in C_k}{n_j}}f_{i}(\omega)
\label{eq:clustering}
\end{equation}
where $\sum_{j\in C_k}{n_j}$ corresponds to the samples-size's sum  of all clients inside cluster $C_k$ and :

\begin{equation}
f_{i}(\omega)=\mathbb{E}_{(X,Y) \sim D_i}[L_{i}(X,Y,\omega)] \hspace{3pt} \forall \hspace{1pt} i \hspace{1pt} \in C_k
\end{equation}
that is the expected value of the local loss function $L_i$ calculated with feature and target $(X,Y)$ following the distribution of dataset $D_i$ with model of parameters $\omega\in \mathbb{R}^d$. Once the clusters are formed, each cluster can independently run a standard FL protocol (e.g., FedAvg) restricted to its members, thereby producing $K$ federated models, each tailored to the data distribution of its cluster.

\begin{figure}[htbp]
    \centering
    \begin{subfigure}[b]{0.48\textwidth}
        \centering
        \includegraphics[width=1.05\linewidth]{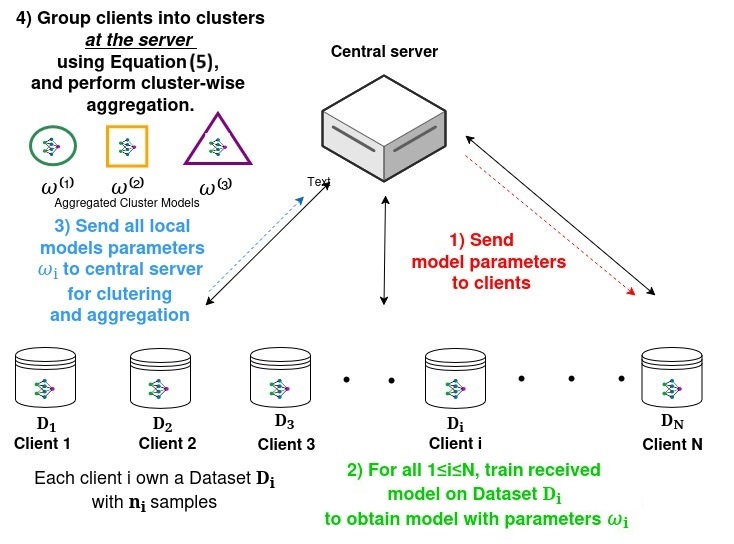}
        \caption{Schematic of Server-side CFL. Clustering is done by the Server using clients' model parameters.}
        \label{fig:cfl_server}
    \end{subfigure}
    \hfill 
    \begin{subfigure}[b]{0.5\textwidth}
        \centering
        \includegraphics[width=\linewidth]{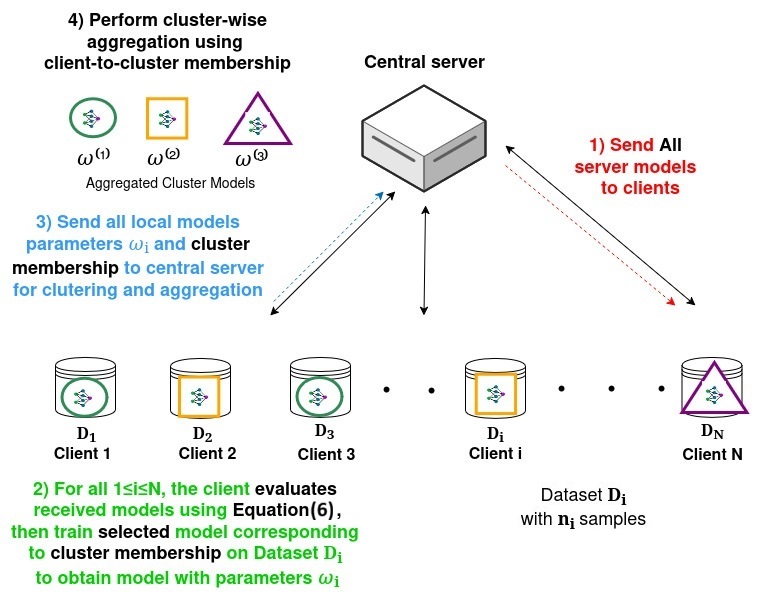}
        \caption{Schematic of Client-side CFL. Clients calculated their cluster membership locally.}
        \label{fig:cfl_client}
    \end{subfigure}
    \begin{subfigure}[b]{0.52\textwidth}
        \centering
        \includegraphics[width=\linewidth]{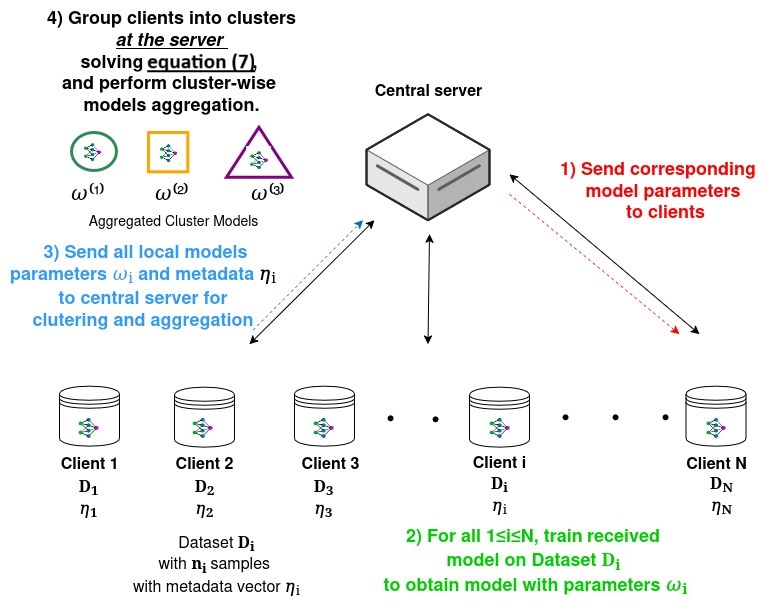}
        \caption{Schematic of Metadata-based CFL. Clients share local metadata with the server, and the server uses collected metadata to perform clustering.}
        \label{fig:cfl_metadata}
    \end{subfigure}
    \caption{Comparison of different CFL approaches from taxonomy. In each figure, steps 1 to 4 consist of a full communication round of CFL and can be repeated as needed.} 
    \label{fig:global_CFL}
\end{figure}

Both the identification of cluster memberships $C_k$ and the optimization of Equation~\ref{eq:clustering} must be carried out while respecting the original assumptions of FL, where only model weights (or gradients) and dataset sizes are shared. However, it is important to note that some papers \citep{ppcfl,CFLGT} relax this assumption by allowing the exchange of additional encrypted metadata. Based on this, a taxonomy is proposed to classify CFL solutions found in the literature, namely: 
\begin{itemize}
    \item \textit{Server-side CFL} (Section~\ref{sec:server-side}), where clustering is performed solely using shared model weights or gradients (Figure~\ref{fig:cfl_server}).
    \item \textit{Client-side CFL} (Section~\ref{sec:client-side}), where clustering membership is determined locally by the clients using client training loss (Figure~\ref{fig:cfl_client})
    \item \textit{Metadata-based CFL} (Section~\ref{sec:metadata-based}), which allows the sharing of additional encrypted metadata with the server for clustering (Figure~\ref{fig:cfl_metadata}).
\end{itemize}

\subsubsection{Server-side CFL}\label{sec:server-side}


The initial approach, referred in this paper as Server-side CFL (see Figure~\ref{fig:cfl_server}), is used in \citep{ghosh2019robust, sattler2020clustered, briggs2020federated, geneticcfl, FedEM} and proposes to solve Equation~\ref{eq:clustering} by only using model weights or gradient information. This strategy has the advantage of requiring minimal additional setup compared to the baseline FL approach. In the most straightforward version, the central server calculates clusters by applying a k-means algorithm with $\ell_2$ (Ghosh et al.~\cite{ghosh2019robust}) to the locally trained model weights, after which cluster-specific federated models are optimized separately for each cluster using the desired FL protocol. 

To formalize this, we introduce for each cluster $C_k$ a representative point $\mu_k \in \mathbb{R}^d$, which serves as the cluster centroid in the model-parameter 
space (i.e., a model representative of the cluster). The central server assigns each client $i \in \{1,\ldots,N\}$ to a cluster by selecting the representative point 
closest to its local model parameters $\omega_i \in \mathbb{R}^d$. This clustering step can be expressed as the following k-means--like optimization problem:
\begin{equation}
\underset{\{\mu_k\}_{k=1}^K}{\arg\min}{\sum_{k=1}^{K}\sum_{i=1}^{N}\mathbb{1}_{i \in C_k} \hspace{1pt} \text{dist}(\omega_{i},\mu_{k})} 
\label{eq:clustergrad}
\end{equation}

This approach has been adopted in many studies, with $\textbf{dist}$ denoting a generic distance function applied to model parameters or their gradients, and $\mathbb{1}_{i \in C_k}=1$ if client $i$ belongs to cluster $C_k$, and $0$ otherwise. The distance can also be replaced by a similarity metric, in which case Equation~\ref{eq:clustergrad} becomes a maximization problem. Early server-side CFL methods, such as those by Sattler et al.~\citep{sattler2020clustered} (MTCFL), relied on cosine similarity within an iterative bipartitioning procedure, where the most similar pairs are grouped until two coherent subsets naturally emerge, and this process is repeated recursively to find clusters.

A central assumption behind these approaches is that locally trained model weights implicitly encode each client’s data distribution. MTCFL~\citep{sattler2020clustered} formalizes this idea by treating model updates as surrogates for data heterogeneity and provides a theoretical guarantee that a meaningful bipartition exists when multiple underlying data distributions are present. In this view, model parameters act as high-level summaries of local data, enabling the server to infer clients' similarity without accessing raw data. These works represent the pioneering approaches of CFL. Since the central server determines the cluster assignment, we refer to this family of methods as \textbf{server-side} throughout the remainder of this paper.

\subsubsection{Client-side CFL}\label{sec:client-side}

As an alternative, several studies delegated determination of cluster membership to the clients (see Figure~\ref{fig:cfl_client})\citep{IFCA,mansour2020three,flsc,ruan2022fedsoft,gong2022adaptive,long2023multi}. In these approaches, clients are presented with a set of candidate cluster representative models, and each decides locally—using its data—which matches best. In the original method IFCA by Ghosh et al.~\citep{IFCA}, this assignment is performed by selecting the representative model that minimizes the client’s training loss.

Formally, let $\mu_{k} \in \mathbb{R}^{d}$ denote the representative model of cluster $C_k$ for all $k \in {1,\ldots,K}$, which plays a central role in cluster formation. In \citep{IFCA}, these representatives are randomly initialized by the server. Each client $i \in {1,\ldots,N}$ then evaluates all candidates $\mu_k$ on its local dataset and selects the one that yields the lowest training loss:
\begin{equation}
\omega_i =
\underset{\{\mu_k\}_{k=1}^K}{\arg\min}
\,\mathbb{E}_{(X,Y) \sim D_i}\left[L_i(X,Y,\mu_k)\right],
\quad \forall i \in {1,\ldots,N}.
\label{eq:training_loss}
\end{equation}

Each client assigns itself to the $\mu_k$ that minimizes the local loss $L_i$ over all samples of dataset $D_i$. After each client selects a cluster, the representatives are updated at each communication round by aggregating the clients' trained models within the corresponding cluster \citep{IFCA}. This process can be repeated iteratively until cluster assignments stabilize.

The main advantage of this approach is that it leverages the distributed nature of FL for cluster assignment. As stated by Ghosh et al.~\citep{IFCA}, “one of the major advantages of [their] algorithm is that it does not require a centralized clustering algorithm, and thus significantly reduces the computational cost at the center machine.”

However,  it also presents drawbacks. As argued in \citep{long2023multi}, these methods introduce higher communication and computation overhead, since clients must download and evaluate multiple global models during the assignment step. This additional burden can become problematic for resource-constrained clients.

Since clients determine the cluster assignment, we call those methods client-side CFL. For clarity, they refer strictly to methods where the clustering computation is carried out entirely on the client side, with no clustering operation performed on the server. 

\subsubsection{Metadata-based CFL} \label{sec:metadata-based} 

The third category (see Figure \ref{fig:cfl_metadata}) considered in this study relies on sharing additional clients' metadata to the server for clustering purposes \citep{dennis2021heterogeneity, MD-ICFL, PACFL, FedPEC, CFLGT, ppcfl}. For each client $i \in \{1,\ldots,N\}$, we define a metadata vector $\eta_i$, which may include statistical summaries of the local dataset (e.g., label distributions, feature statistics, or centroids of local data clusters). For each cluster $C_k$, we then compute a representative point $\delta_k$ (typically the centroid of metadata vectors) for all $k \in \{1,\ldots,K\}$. Clients membership is determined through a formulation analogous to Equation~(\ref{eq:clustergrad}):

\begin{equation}
\underset{\{\delta_k\}_{k=1}^K}{\arg\min}{\sum_{k=1}^{K}\sum_{i=1}^{N}\mathbb{1}_{i \in C_k}\,\text{dist}(\eta_i,\delta_k)}.
\label{eq:metadata}
\end{equation}

Here, \textbf{dist} denotes a generic distance between metadata vectors, and $\mathbb{1}_{i \in C_k}=1$ if client $i$ is assigned to cluster $C_k$, and 0 otherwise. Since the cluster assignment is driven by metadata, we refer to these approaches as \textit{metadata-based}. Although Equation~\eqref{eq:metadata} is expressed using a distance, an equivalent formulation can be obtained with a similarity function, with the clustering objective becoming a maximization problem.

In terms of design, metadata-based CFL remains structurally similar to server-side approach, as the server still performs the grouping. The key distinction lies in replacing or supplementing model weights, updates, or gradients with metadata explicitly provided by clients. When well-chosen, such metadata can substantially improve clustering quality and reduce computational cost, as metadata vectors are typically far lower-dimensional than model parameters. Another practical advantage is that, unlike model weights or gradients that evolve at every round, metadata can often remain static, which reduces communication overhead since clients do not need to transmit them at every iteration.

However, these benefits come with weaker privacy guarantees, especially when the shared metadata consists of clients’ dataset statistics, which reveal sensitive information. Even when protected using security mechanisms such as differential privacy~\cite{ppcfl}, metadata introduces potential privacy and security risks~\cite{kairouz2021advances}. Such trade-offs depend on the application domain and whether users can explicitly consent to metadata sharing. In sensitive contexts, even aggregated statistics may be infeasible to share due to regulatory or ethical constraints, limiting the applicability of such approaches.

Early examples include Dennis et al.~\cite{dennis2021heterogeneity}, who proposed clustering data locally and sharing the resulting cluster centroids as metadata, and Zhang et al.~\cite{MD-ICFL}, who combined local label-distribution vectors with model distances. These approaches highlight how leveraging informative, low-dimensional metadata can significantly improve the efficiency and quality of clustering.

\subsubsection{Summary of Strength and Weakness of CFL Categories} \label{sec:summary-cfl-solutions}

The comparison between the three CFL categories illustrates that each approach offers a different balance between privacy, computational cost, and clustering effectiveness. \textbf{Server-side} approaches align most closely with the original FL assumptions, preserving strong privacy and imposing no additional computation burden on clients. However, centralized clustering on high-dimensional model weights is computationally expensive, leading to scalability issues with large models or many clients. \textbf{Client-side} methods also preserve strong privacy while distributing the clustering computation across clients, reducing server computational burden. They impose a higher burden on resource-constrained devices and may increase communication overhead. \textbf{Metadata-based} solutions enhance clustering efficiency and effectiveness by operating on lightweight descriptors, reducing clustering costs compared to using model weights. Additionally, unlike model parameters, statistical descriptors are often invariant over time, ensuring stable cluster formation metrics 
across communication rounds. This efficiency, however, comes at the cost of 
weaker privacy guarantees, as sharing metadata exceeds the strict constraints 
of the standard FL exchange protocol.

\begin{table}[htbp]
\centering
\footnotesize
\begin{tabular}{|p{2.6cm}|c|c|c|c|c|}
\hline
\textbf{CFL Category} 
& \textbf{Privacy} 
& \shortstack{\textbf{Server}\\\textbf{Computation}\\\textbf{Burden}} 
& \shortstack{\textbf{Client}\\\textbf{Computation}\\\textbf{Burden}} 
& \shortstack{\textbf{Communication}\\\textbf{Efficiency}} 
& \shortstack{\textbf{Cluster}\\\textbf{metrics}\\\textbf{Stability}}\\
\hline

\textbf{Server-side} 
& $++$ 
& $++$ 
& $-$ 
& $+$ 
& $-$ \\
\hline

\textbf{Client-side} 
& $++$ 
& $-$ 
& $++$ 
& $-$ 
& $-$ \\
\hline

\textbf{Metadata-based} 
& $-$ 
& $+$ 
& $-$ 
& $++$ 
& $++$ \\
\hline

\end{tabular}
\caption{Expanded comparison of CFL categories across privacy, computation burden, communication efficiency, and metrics stability.}
\label{tab:cfl_extended}
\end{table}

The choice of an approach thus depends heavily on the application domain and the acceptable trade-off between privacy, computational efficiency, and cluster formation metrics. To summarize these contrasting criteria, Table \ref{tab:cfl_extended} provides a compact view of the relative performance of each CFL category across five high-level metrics.

\subsection{Review and Categorization of core CFL Solutions} \label{sec:clf_classification}
\begin{figure}[htbp]
    \adjustbox{width=0.8\textwidth}{\includegraphics{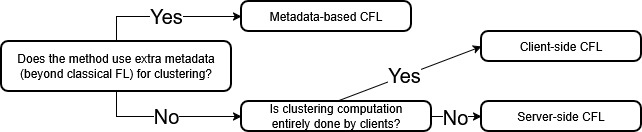}}
    \caption{Taxonomy Diagram for classification of CFL methods.}
    \label{fig:CFL_classif_diag}
\end{figure}

In this section, we focus exclusively on \textbf{core CFL approaches that are domain agnostic}—that is, algorithms designed to handle non-IID data without being tied to any task, model architecture, or application domain. All selected articles follow the guidelines described in Section~\ref{sec:methodology}. Our objective is to provide a classification of those CFL approaches according to our taxonomy. To clarify this classification, we propose a diagram in figure~\ref{fig:CFL_classif_diag}. First, it determines whether clustering relies on metadata shared with the server beyond standard FL assumptions, and secondly, whether the clustering computation is performed entirely by clients or involves the server.

\begin{table}[ht!]
\centering
\caption{Chronological timeline of CFL approaches with clustering methods (server-side, client-side, or metadata-based) and corresponding evaluation metrics.}
\label{tab:CFL_classification}
\tiny
\begin{tabular}{|p{2.2cm}|p{0.85cm}|p{2.2cm}|p{2.5cm}|p{1cm}|p{3.5cm}|}
\hline
\textbf{Reference} & \textbf{Algorithm} & \multicolumn{3}{c|}{\textbf{CFL Category}} & \textbf{Metrics} \\ \cline{3-5}
 &  & \textbf{Server-side} & \textbf{Client-side} & \textbf{Metadata-based} &  \\ \hline

Ghosh et al. 2019~\cite{ghosh2019robust}\footnote{Foundational CFL papers.} & N/A & Trimmed k-means & \graycell & \graycell & $L_2$ norm of models weights\\ \hline
Sattler et al. 2020~\cite{sattler2020clustered}\footnotemark[2] & MTCFL & Iterative bi-partitioning & \graycell & \graycell & Cosine similarity of models gradients \\ \hline
Mansour et al. 2020~\cite{mansour2020three}\footnotemark[2] & HypCluster & \graycell & Cluster selection by minimum & \graycell & Local training loss  \\ \hline
Briggs et al. 2020~\cite{briggs2020federated}\footnotemark[2] & FL+HC & Hierarchical clustering \textbf{(HC)} & \graycell & \graycell & $L_1$/$L_2$/Cosine similarity of models update\\ \hline
Ghosh et al. 2020~\cite{IFCA}\footnotemark[2] & IFCA & \graycell & Cluster selection by minimum & \graycell & Local training loss \\ \hline
Dennis et al. 2021\cite{dennis2021heterogeneity} & k-Fed & \graycell & \graycell & K-means & $L_2$ over local clients data centroids \\ \hline
Li et al. 2021\cite{flsc} & FLSC & \graycell & Cluster selection by minimum & \graycell & Local training loss \\ \hline 
Agrawal et al. 2021\cite{geneticcfl} & GeneticCFL & \graycell & \graycell & DBSCAN & clients learning rate/batch-size \\ \hline
Marfoq et al. 2021\cite{FedEM} & FedEM & \graycell & Expectation-Maximization & \graycell & Local samples contribution for each model \\ \hline
Duan et al. 2021\cite{duan2021flexible} & FlexCFL & HC/K-means & \graycell & \graycell & MADD-based metric on models update \\ \hline
Jamali-Rad et al. 2022\cite{FLT} & FLT & \graycell & \graycell & HC & Centroids of Local Autoencoder outputs\\ \hline
Gong et al. 2022\cite{gong2022adaptive} & AutoCFL & \graycell & Voting scheme & \graycell & $L_2$ Model similarity matrix shared with clients\\ \hline
Ruan et al. 2022\cite{ruan2022fedsoft} & FedSoft & \graycell & Cluster selection by minimum & \graycell & Local loss per-sample score \\ \hline
Long et al. 2023\cite{long2023multi} & FeSEM & \graycell & Cluster selection by minimum & \graycell & Local model distance to cluster models \\ \hline
Jothimurugesa al 2023\cite{FedDrift} & FedDrift & \graycell & Cluster selection by minimum & \graycell & Local training loss \\ \hline
Jingyu et al. 2023\cite{ascfl} & ASCFL & \graycell & \graycell & K-means & Custom metric over clients loss and weights \\ \hline
Yihan et al. 2023\cite{ICFL} & ICFL & Iterative bipartitioning & \graycell & \graycell & SVD of cosine similarity matrix of models weights \\ \hline
Tun et al. 2023\cite{CP-CFL} & CP-CFL & \graycell & Cluster selection by minimum & \graycell & Loss \\ \hline
Vahidian et al. 2023\cite{PACFL} & PACFL & \graycell & \graycell & HC & Principal angle from SVD-reduced features \\ \hline
Cai et al. 2023\cite{FedCE} & FedCE & \graycell & Cluster selection by minimum & \graycell & Historical loss score (Euclidean) \\ \hline
Morafah et al. 2023\cite{FLIS} & FLIS & Hierarchical Clustering & \graycell & \graycell & Cosine similarity on inferences over public dataset \\ \hline
Ma et al. 2023\cite{FedCAM} & FedCAM & \graycell & Cluster selection by minimum & \graycell & Same as FeSEM or IFCA \\ \hline
Liang et al. 2023\cite{liang2023efficient} & FedEOC & K-means & \graycell & \graycell & Cosine similarity of model weights \\ \hline
Mehta et al. 2023\cite{mehta2023greedy} & FLACC & HC & \graycell & \graycell & Cosine similarity of model weights  \\ \hline
Yu et al. 2024\cite{FPFC} & FPFC & Hard Thresholding & \graycell & \graycell & Penalized Euclidean distance of model weights \\ \hline
Zhang et al. 2024\cite{MD-ICFL} & MD-ICFL & \graycell & \graycell & Min Distance & L1 norm over weighted softmax outputs \\ \hline
Gao et al. 2024\cite{FedPEC} & FedPEC & \graycell & \graycell & GMM & Clients local Shapley values \\ \hline
Zhang et al. 2024\cite{FedGK} & FedGK & HC & \graycell & \graycell & Cosine similarity of weight updates \\ \hline
Liu et al. 2024\cite{CFLGT} & CFLGT & Affinity Propagation & \graycell &  \graycell & Gradient trajectories (pull/push forces) \\ \hline
Guo et al. 2024\cite{FedRC} & FedRC & \graycell & Bi-level optimization & \graycell & Likelihood-weighted importance + loss \\ \hline
Luo et al. 2024\cite{ppcfl} & pp-CFL & \graycell & \graycell & COPRA &  Noisy label distribution \\ \hline
Gong et al. 2024\cite{HiCFL} & HiCFL & Iterative Bi-partitioning & \graycell & \graycell & Cumulated update cosine similarity per rounds \\ \hline
Ren et al. 2025\cite{TDCFL} & TDCFL & \graycell & \graycell & HC & Distance of SVD decomposition of data \\ \hline
Cheng et al. 2025\cite{SnapCFL} & SnapCFL & K-Medoids / HC / DBSCAN & \graycell & \graycell & Classifier two-sample test \\ \hline
Zeng et al. 2025\cite{StoCFL}& StoCFL & Agglomerative Clustering & \graycell & \graycell & Frozen model's Gradient cosine similarity \\  \hline
Guo et al. 2025\cite{HCFL+} & HCFL+ & \graycell & Bi-level optimization & \graycell & Custom loss function \\ \hline
\end{tabular}
\end{table}
\normalsize
We provide a high-level synthesis of reviewed papers in Table~\ref{tab:CFL_classification}, ordered by publication date. The column “Algorithm” lists the acronym of the approach, either as defined in the original paper or adopted in subsequent works. “CFL Category” summarizes the procedure used for clustering and assigns each method to a category within the taxonomy introduced in Section~\ref{sec:cfltaxonomy}. In server-side and metadata-based approaches, it typically refers to a classical clustering algorithm, whereas in client-side methods, it describes how clients determine their cluster membership. The “Metric” column specifies the criterion for similarity between clients in server-side, which determines cluster membership in client-side CFL, and which metadata is used in metadata-based. It is worth noting that several algorithms share similar clustering mechanisms or metrics in Table~\ref{tab:CFL_classification}, yet differ in other aspects of the CFL mechanisms discussed in Sections \ref{sec:scalable} to \ref{sec:security_concerns}. To accurately capture the historical development of CFL, foundational studies are classified in chronological order, according to the date of publication of their first preprint release, as many were cited before formal publication, while later works follow their official publication date. This chronological organization situates the evolution of CFL within a unified timeline, while providing a taxonomy-based categorization and highlighting the key mechanisms of each method.

\underline{The earliest theoretical formulation of CFL} we identified is by \underline{Ghosh et al.~\cite{ghosh2019robust}} in 2019 with a server-side approach. Their approach tackles a non-IID FL setting in the presence of adversarial machines by applying variants of k-means clustering on locally trained models to form clusters during the initial FL round, followed by robust optimization within each cluster to perform federated learning. The first explicit use of the term CFL appears in Sattler et al.~\cite{sattler2020clustered} in 2020. The authors formalized CFL as a model-agnostic framework where clients are partitioned into groups based on the cosine similarity of their model updates. The server builds a similarity matrix and applies an iterative bipartitioning process that recursively splits clusters whenever their internal similarity falls below a predefined threshold. Within each cluster, the algorithm merges the most similar client pairs until only two groups remain, forming the partitions. Each resulting cluster is then treated as an independent FL instance, and the partitioning is re-evaluated at every communication round. This approach is often referred to simply as MTCFL (Multi-Task CFL), as each cluster model can be viewed as an independent task. Together, these two early methods established the foundations of server-side CFL. 

Later in 2020, HypCluster~\citep{mansour2020three}  and IFCA~\citep{IFCA}, both introduced similar ideas, laying the foundation of client-side CFL. Both propose to clients multiple models which locally select the model that minimizes its training loss (As explained in section~\ref{sec:client-side}). Both papers also introduce theoretical advances for CFL: HypCluster provides a generalization bound showing that learning multiple cluster-specific models yields a provable trade-off between global generalization and local specialization, whereas IFCA establishes a convergence-rate analysis for its algorithm. However, a key limitation of these methods is their sensitivity to initialization. Poorly chosen initial models may lead to empty clusters, sometimes requiring multiple runs to obtain stable partitions.

In the same year, Briggs et al. \cite{briggs2020federated} proposed FL+HC which proposes the use of  \textbf{Hierarchical Clustering (HC)} with model update similarity. Different similarities ($L_1$, $L_2$, cosine) and linkage (rule to measure inter-cluster distance) were evaluated. Instead of iteratively calculating clusters at each communication round, FL+HC proposes to apply a one-shot clustering after a prior predefined number of standard FL rounds, making it more computationally efficient. Another major contribution of this paper is \underline{the first explicit use of a non-IID data taxonomy} to create an evaluation benchmark for CFL, something we will discuss more thoroughly in Section~\ref{sec:cfl_eval}. The five aforementioned papers have given a solid foundation for CFL approach to build upon and remain relevant even today. All following papers in the timeline will have at least an experimental comparison to one of these previous CFL articles (see Figure~\ref{fig:cfl_timeline} that we will discuss in Section~\ref{sec:cfl_eval}).  

The \underline{first domain-agnostic metadata-based CFL approach} was introduced by Dennis et al.~\citep{dennis2021heterogeneity} in 2021 with k-FED, which achieves competitive results to IFCA. In this method, each client performs local k-means clustering on their dataset and sends the resulting centroids to the server, which are interpreted as metadata summarizing local data distributions. The server then applies k-means on those centroids to form groups of clients. Agrawal et al.~\cite{agrawal2021genetic} proposed GeneticCFL, a metadata-based CFL algorithm that clusters clients according to local hyperparameters such as learning rate and batch size. The authors’ intuition is that optimal hyperparameters reflect characteristics of data distributions. Clients are first assigned several candidate hyperparameter configurations and select the one that best fits their data. Over time, these hyperparameters evolve through genetic optimization, and the server clusters clients using DBSCAN computed from their Euclidean distances. 

From this point onward, the overall framework of CFL becomes more stable, and research gradually shifts from proposing entirely new paradigms to refining and extending existing ones. Most subsequent studies in Table~\ref{tab:CFL_classification} can be seen as evolutions of the foundational IFCA for client-side approaches and MTCFL for server-side ones, or as adaptations of MTCFL that incorporate metadata in the case of metadata-based CFL. These works build on earlier principles to address broader challenges of CFL. Their goals include determining the appropriate number of clusters, improving computational and communication efficiency, introducing soft client assignments or local adaptation mechanisms to better handle diverse non-IID scenarios, and enhancing adaptability to dynamic environments such as newcomer arrival or data drift. Rather than redefining the CFL paradigm, these later contributions focus on making it more practical, flexible, and robust.

\subsubsection{Number of Clusters}\label{sec:number_of_clusters}
Like most clustering algorithms, the earliest CFL methods required the number of clusters $k$ to be predefined by the user. In practice, $k$ is often chosen based on domain knowledge or empirically by selecting the value that yields the best performance on the FL task. However, since the number of underlying client data distributions is unobservable, determining this value is challenging and incurs a substantial communication cost. Server-side and metadata-based algorithms benefit from decades of clustering research, as they can directly apply standard clustering techniques that do not require specifying the number of clusters beforehand. Several CFL methods employ algorithms such as HC \citep{briggs2020federated,PACFL,SnapCFL}, iterative bipartitioning \citep{sattler2020clustered,ICFL}, or DBSCAN \citep{geneticcfl}, which implicitly determine the number of clusters using alternative hyperparameters. However, client-side all relies on a fixed number of clusters except HCFL+~\cite{HCFL+}, which inputs an initial number of clusters, then uses metrics thresholding to divide or merge clusters. 

Nevertheless, all CFL algorithms depend on hyperparameters that influence the number of clusters either directly or indirectly. Thus, although recent methods have reduced reliance on a fixed $k$, fully automatic and stable cluster determination in CFL remains an open research challenge.

\subsubsection{Scalability and Efficiency Enhancement Strategies}\label{sec:scalable}

Several works focus on improving scalability and efficiency, addressing both computational and communication costs. The problem becomes increasingly severe as the number of clients and model parameters grows. Computing pairwise similarities and performing clustering quickly becomes resource-intensive. These strategies aim to reduce the computational burden, accelerate convergence, and simplify client participation.

It is interesting to note that early CFL approaches such as FL+HC~\cite{briggs2020federated} and k-FED~\cite{dennis2021heterogeneity} proposed one-shot clustering after several rounds of standard FL, whereas most CFL methods are iterative. This approach is computationally efficient since clustering is performed only once. However, in non-metadata-based settings, identifying optimal clusters in a one-shot manner is challenging because the features used for clustering, such as local model weights or losses, evolve across rounds.

In server-side CFL, scalability issues often arise because clustering is performed directly on model, which may contain millions of parameters. To address this \textbf{high-dimensional low-sample-size (HDLSS)} challenge, Duan et al. proposed FlexCFL~\cite{duan2021flexible}. It adopts MADD-based similarity metrics tailored for HDLSS data. Additionally, truncated \textbf{singular value decomposition (SVD)} reduces the dimensionality of the similarity matrix, significantly lowering computational complexity. Similarly, ICFL~\cite{ICFL} applies the same decomposition and iterative bipartitioning to improve clustering.

Other approaches focus on computational speed by adjusting clustering dynamics. For instance, Mehta et al. introduced FLACC~\cite{mehta2023greedy}, a greedy clustering strategy that groups clients before full model convergence, reducing training time while maintaining competitive performance. Likewise, FedEOC~\cite{liang2023efficient} improves scalability by comparing models through a single representative layer rather than full model parameters.

Client-side CFL also integrates lightweight mechanisms to enhance efficiency. AutoCFL~\cite{gong2022adaptive} reduces computation by calculating inter-model distances using only partial model weights. FeSEM~\cite{long2023multi} adopts layer-wise matching aggregation, inspired by~\cite{wang2020federated}, to achieve faster convergence and lower training overhead.

Finally, some metadata-based CFL methods improve scalability by operating in a reduced feature space. Examples include ASCFL~\cite{ascfl}, CP-CFL~\cite{CP-CFL}, and PACFL~\cite{PACFL}, which rely on server-trained or pretrained autoencoders (often on public datasets) to minimize local computational costs while preserving meaningful data representations. As they extract low-dimensional embeddings from client datasets through autoencoders separate from the task model. However, training such models can still be costly from a CFL perspective.

\subsubsection{Soft clustering}\label{sec:soft}
CFL assumes that devices can be divided into distinct data distributions. Unfortunately, real-world data are often more complex. In some cases, each client’s dataset may originate from a mixture of overlapping distributions rather than a single one. Interestingly, none of the server-side approaches studied uses \textbf{soft clustering (SC)}. The only metadata-based SC approach is FLT~\cite{FLT}, uses a soft HC algorithm. However, the use of SC is technically applicable in both approaches using standard SC algorithms. 

For client-side, Li et al.~\cite{flsc} highlight that algorithms like IFCA assume non-overlapping clusters, which leads to inefficient utilization of local information, since a client’s knowledge contributes to only one cluster per round. They propose FLSC, a SC version of IFCA. Locally, each client selects several of the best-performing models instead of just one, averages them, and trains the resulting model before sending it back to the server, thereby improving results over IFCA in their evaluations.

FedEM~\cite{FedEM} introduces an approach based on the \textbf{Expectation–Maximization (EM)} algorithm. The server maintains multiple global models and broadcasts them to all clients. Each client is associated with all clusters simultaneously and, during the local EM step, computes the contribution of each data sample to each cluster model. These contributions are then used to weight local updates for all models before being sent back to the server, which aggregates them to refine each global model. 
Similarly, several other algorithms~\citep{ruan2022fedsoft, FedCE, FLIS, FedRC, ppcfl, HCFL+} rely on client-derived cluster importance scores to simulate soft clustering.

\subsubsection{CFL Dynamics: Client Selection, Newcomers, Distribution Drift, and Re-Clustering}\label{sec:newcomers_dynamic}

Early CFL strategies, such as Sattler et al. MTCFL~\cite{sattler2020clustered} assumes full client participation during clustering rounds. However, FL inherently operates in a dynamic environment, making this assumption often unrealistic. Several studies~\cite{IFCA, duan2021flexible, FLT, gong2022adaptive, ICFL, mehta2023greedy} incorporated and evaluated random client selection in their designs. For example, AutoCFL~\cite{gong2022adaptive} and ICFL~\cite{ICFL} propose incremental clustering strategies that preserve past models and cluster memberships of inactive clients, allowing clusters to evolve progressively.

Another important scenario, distinct from standard client selection, is the appearance of newcomers. A newcomer is a client who joins late or even near the end of training. Fortunately, MTCFL has already proposed a newcomer strategy. Since it relies on iterative bipartitioning, the complete division tree of clients is preserved, allowing newcomers to traverse the tree and find their most suitable cluster. Newcomer strategies generally aim to assign new clients to existing clusters. Server-side approaches such as FlexCFL~\citep{duan2021fedgroup} assign newcomers by evaluating their similarity to existing cluster aggregates. FlexCFL also includes a client migration mechanism to handle data distribution drift, using the Wasserstein distance between previous and current local distributions to detect significant shifts and treat affected clients as newcomers. Metadata-based methods such as pp-CFL~\cite{ppcfl} and TDCFL~\cite{TDCFL} instead assign newcomers to clusters based on data feature similarity.

Jothimurugesan et al.~\cite{FedDrift} demonstrated that client-side methods can naturally extend their local cluster-selection mechanisms to handle newcomers dynamically. However, this becomes more challenging when newcomers exhibit novel data distributions. A similar issue arises when existing clients experience data distribution drift, requiring re-clustering. Such clients can be treated as newcomers or may form a new cluster that is later made available to other clients in a typical client-side fashion.

Finally, features used for clustering in both server-side and client-side approaches, such as model weights, gradients, or losses, are not static between communication rounds. This makes CFL fundamentally different from standard clustering, which operates on fixed data points. One consequence is the emergence of misclustered clients or incorrect cluster counts—clients may end up in unsuitable clusters, or clusters themselves may need to be split or merged. In such cases, migration mechanisms like those in FlexCFL~\citep{duan2021fedgroup} and FedDrift~\cite{FedDrift} allow clients to move between clusters. Furthermore, Guo et al. (2025)~\cite{HCFL+} proposed HCFL+, a multi-tier generalized CFL architecture encompassing various dynamic re-clustering mechanisms. Proposed approaches can be summarized by highlighting that clusters with large intra-cluster distances can be split, as in MTCFL and ICFL~\cite{ICFL}, while those with small inter-cluster distances can be merged, as demonstrated in StoCFL~\cite{StoCFL}.
 
\subsubsection{Local Optimization, Regularization and Personalization} \label{sec:local_optimization}
As aforementioned, CFL aims to identify and group clients originating from similar probabilistic data distributions. However, clients within the same cluster are rarely perfectly homogeneous, as they are subject to various influences. This lack of intra-cluster homogeneity can still affect training stability and local model adaptation. To mitigate such issues, several studies have integrated local optimization, regularization, and personalization strategies directly into CFL design.

In Gong et al. AutoCFL~\citep{gong2022adaptive}, the authors observe that such heterogeneity may arise from varying dataset sizes. They propose a local training adjustment mechanism that adapts the number of training epochs according to dataset size. Clients with larger datasets perform more local updates, while those with smaller train for fewer epochs to accelerate convergence.

Similarly, FeSEM~\citep{long2023multi} extends IFCA by redefining the assignment of clients to clusters based on the distance between local and cluster models, rather than on training loss. It introduces a personalized local objective function with a regularization term that penalizes deviations from the assigned cluster model, thereby balancing personalization and shared cluster learning. Additionally, FedCAM~\citep{FedCAM} introduces an additive modeling structure for IFCA and FESEM in which each client’s model is decomposed into a global component, a cluster-specific component, and a personalized residual. FedCE~\citep{FedCE} adopts a similar personalization approach, where the local aggregation of the cluster models to which a client belongs is added as a regularization component.

FedEOC~\cite{liang2023efficient} adopts a local optimization strategy where each client splits its local dataset into a support set and a query set. The client first updates its model parameters using the support set, then computes new gradients on the query set using the adapted model, which are sent to the server for clustering. This allows the global model to form efficient one-shot clustering quickly. FPFC~\cite{FPFC} reformulates CFL as an optimization problem (ADMM~\cite{han2022survey} scheme with a penalty regularizes pairwise model distances). This mechanism acts as a form of regularized local optimization that balances model fusion and personalization.

More recently, CFL frameworks such as FedRC~\citep{FedRC} and HCFL+~\citep{HCFL+} formulate clustering and optimization as a unified bi-level problem. These SC CFL methods jointly optimize locally cluster importance score (section~\ref{sec:soft}) and model parameters through regularization. These approaches bridge the gap between clustered and personalized FL, showing that personalization and regularization can enhance intra-cluster stability and robustness.

\subsubsection{Security Concerns} \label{sec:security_concerns}
As data privacy lies at the core of FL, ensuring the security of exchanged information remains crucial. Fortunately, since CFL does not alter the underlying FL communication protocol, standard FL security mechanisms can be directly applied~\cite{mothukuri2021survey}. However, only a few core CFL studies reviewed explicitly address security, although some works have explored it.  

Ghosh et al. MTCFL~\cite{ghosh2019robust} already proposed a defense against Byzantine clients in CFL by filtering out outliers by using a trimmed K-means based on the geometric median. Moreover, they demonstrated that rotation-invariant metrics such as cosine similarity or $L_2$ norm can be made compatible with encryption schemes like those introduced by Bonawitz et al.~\cite{bonawitz2017practical}. 

It is worth noting that, as shown in Table~\ref{tab:metadata_risk_simple}, metadata-based CFL methods are inherently more vulnerable to privacy leakage. A promising direction is the integration of differential privacy. Luo et al. introduced pp-CFL~\cite{ppcfl}, which shares clients' differentially private label distribution vectors for clustering, achieving improved privacy. Additionally, Ren et al.  TDCFL~\cite{TDCFL} notes that existing privacy-preserving techniques, such as multiparty secure encryption, homomorphic encryption, and differential privacy, can be further integrated to enhance protection against potential risks. However, secure clustering over encrypted data has been studied (Jaschke et al. \cite{jaschke2018unsupervised}); its adoption remains impractical because performing encrypted clustering introduces high computational cost, making it incompatible with many real-world CFL setups.

\subsubsection{A Privacy-Vulnerability Risk Analysis of Metadata-based CFL}\label{sec:privacy-meta}

As aforementioned, metadata-based CFL relies on low-dimensional descriptors to encode clients' data distribution. While this is communication-efficient and may improve grouping, it also introduces new privacy concerns. Metadata may leak sensitive information or increase vulnerability to inference attacks. With strict confidentiality, sharing metadata may even be prohibited.

We propose to categorize metadata privacy risk in Table~\ref {tab:metadata_risk_simple}. It proposes three risk levels reflecting how much shared metadata exposes the structure of the underlying local dataset and how easily it can be exploited for inference or reconstruction attacks.

\textbf{Level I: Low Risk (Indirect Metadata).} \
Those represent relatively low risk, such as hyperparameter selection or local model losses. These values mainly indicate computational choices or overall model fit, rather than exposing information about the data. 

\textbf{Level II: Significant Risk (Direct Statistical Signatures).} \
Clients share raw statistical descriptors such as label distributions, centroids, or Shapley values. These act as a direct representation of the dataset and can support property-inference attacks, where an adversary tests whether data points or attributes are present in the training data (Melis et al.~\cite{melis2019exploiting}). Furthermore, Shapley values explicitly quantify the contribution of each feature to the model's decision for specific data points. We argue that this could be exploited to infer membership or reveal sensitive information.

In FL, where gradients are already exchanged, releasing such statistics further strengthens privacy leakage. They provide strong information on data, thus significantly improving gradient inversion attacks such as those in Geiping et al.~\cite{geiping2020inverting}.

\textbf{Level III: High Risk (Reduced Feature Representations).} \\
This is the most privacy-sensitive category. Autoencoder embeddings, SVD projections, or other reduced representations of the full client dataset preserve the topological structure of the raw data to ensure quality clustering. However, research demonstrates that these low-dimensional embeddings are highly susceptible to inversion attacks~\cite{dosovitskiy2016inverting}. Those vectors have a risk of being inverted to recover input samples.

\begin{table}[htbp]
\centering
\footnotesize
\caption{Classification of Metadata-based CFL Approaches by Privacy Risk}
\label{tab:metadata_risk_simple}
\begin{tabular}{|p{2.5cm}|p{5.5cm}|p{4.5cm}|}
\hline
\textbf{Risk Level} & \textbf{Shared Metadata Type} & \textbf{Representative Approaches} \\
\hline
\textbf{I. Low} & Local Hyperparameters; Local models training loss & GeneticCFL \cite{geneticcfl}, ASCFL \cite{ascfl} \\
\hline
\textbf{II. Moderate} & Data Centroids; Local Sample Contribution Weights; Shapley Values; Noisy labels distribution vectors & k-Fed \cite{dennis2021heterogeneity}, MD-ICFL \cite{MD-ICFL}, FedPEC \cite{FedPEC}, pp-CFL \cite{ppcfl}\\
\hline
\textbf{III. High} & Autoencoder Feature Embeddings; SVD-based Reduced Features & FLT \cite{FLT}, PACFL \cite{PACFL}, TDCFL \cite{TDCFL} \\
\hline
\end{tabular}
\end{table}

\subsection{Overview of CFL Workflow design}\label{sec:workflow}

\begin{figure}[htbp!]
    \adjustbox{trim=0cm 0cm 0cm 3cm, clip, width=0.7\textwidth}{\includegraphics{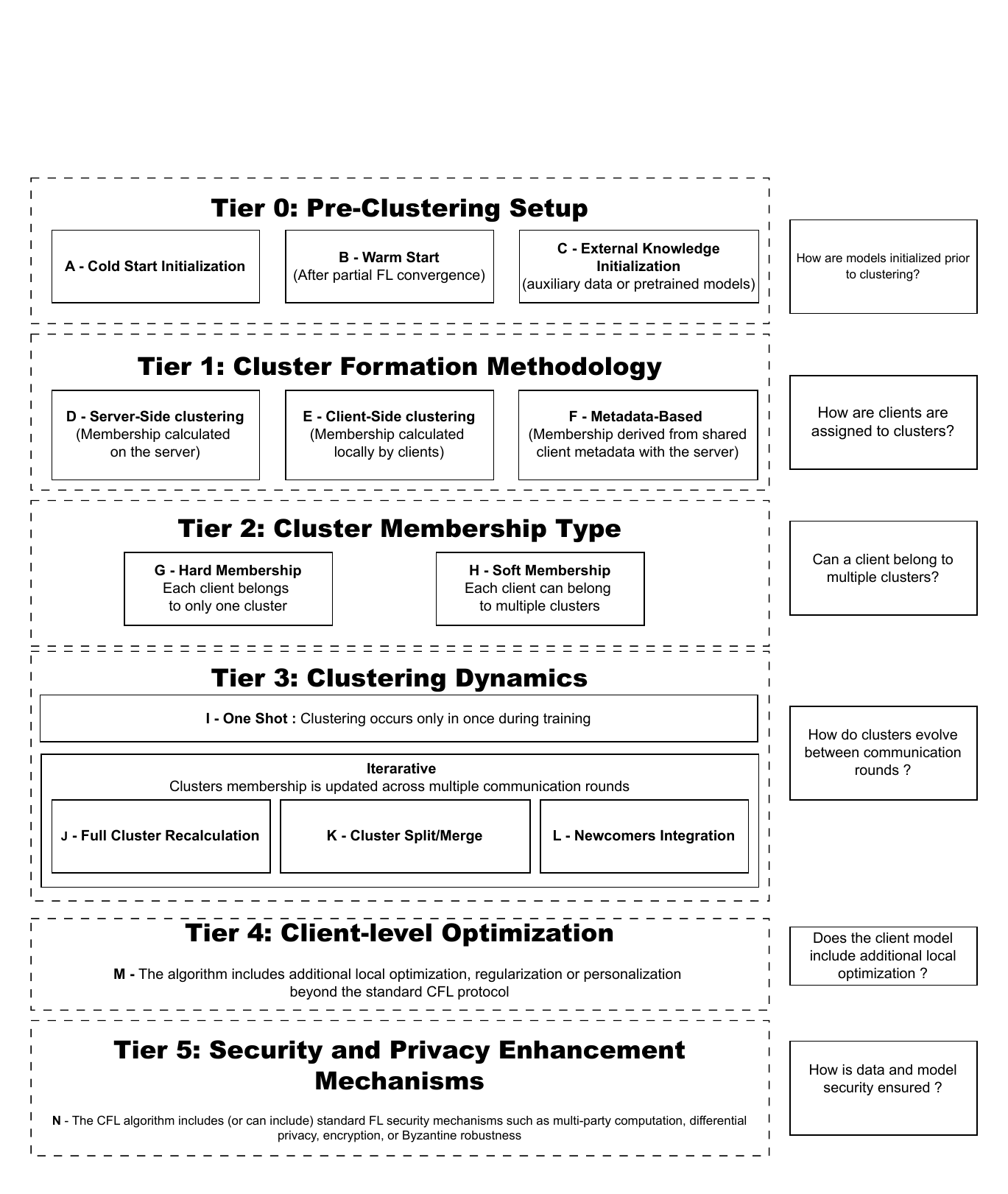}}
    \caption{Generalized CFL System workflow: Each tier captures a specific dimension of the CFL pipeline, enabling structured analysis and comparison of different CFL strategies.}
    \label{fig:FLframework}
\end{figure}
Designing a CFL algorithm involves multiple interacting components that are often presented as a single integrated mechanism in the literature. To motivate a more structured view, Guo et al. (2025)~\cite{HCFL+} propose a multi-tier CFL system that integrates several mechanisms from IFCA, FedEM, and FedRC. Their architecture combines soft clustering with dynamic strategies such as those discussed in Section~\ref{sec:newcomers_dynamic}. 

Intending to provide a unified perspective that encompasses server-side, client-side, and metadata-based CFL approaches for handling non-IID data, and inspired by this architecture, we introduce a generalized CFL workflow that extends their formulation to cover the broader methodological landscape of CFL (see Figure \ref{fig:FLframework}). This workflow is designed to clarify and simplify the key aspects that must be considered when designing or analyzing CFL solutions.

This framework consists of five layers, each divided into one or more tiers, and each tier is denoted by a letter for ease of reference:
\begin{itemize}
\item 
\textbf{Tier 0 addresses the pre-clustering phase.}
A corresponds to a cold start (no specific initialization),
B to warm start (clustering begins after several convergence iterations),
and C to external initialization (for instance, transfer learning or metadata-guided initialization).
\item 
\textbf{Tier 1 defines the global CFL strategy (see Section~\ref{sec:cfltaxonomy}).}
D denotes server-side clustering based on model similarity,
E denotes client-side clustering through local selection among multiple models,
and F denotes metadata-based clustering using external features and similarity metrics.
\item 
\textbf{Tier 2 defines the global CFL strategy (see Section~\ref{sec:soft}).}
G denotes Hard Clustering Membership, meaning each client belongs to only one cluster,
H denotes Soft Clustering Membership, meaning each client can belong to multiple clusters.
\item 
\textbf{Tier 3 concerns clustering dynamics (see Section~\ref{sec:newcomers_dynamic}).}
Tier I corresponds to one-shot clustering,
J to Full  Reclustering,
K to cluster splitting or merging,
and L to newcomer integration.
\item 
\textbf{Tier 4 concerns client-level optimization}, and is denoted by M, some client-level optimization mechanism (see Section~\ref{sec:local_optimization}).
\item 
\textbf{Tier 5 concerns security and privacy.} It is denoted by N. Although such mechanisms are rarely included explicitly in CFL studies, they remain essential to the FL paradigm. As discussed in Section~\ref{sec:security_concerns}, CFL remains fully compatible with standard FL security practices.
\end{itemize}

Table~\ref{tab:CFL_letters} summarizes studied CFL approaches with their corresponding tier classifications; each algorithm is annotated with the letters corresponding to the mechanisms it implements. It complements Table~\ref{tab:CFL_classification} and aims to highlight which challenges each approach specifically addresses or requires further investigation. Unsurprisingly, tiers 4 and 5 are the least studied in the reviewed CFL approaches, as they remain somewhat complementary to the inter-client separate data distribution, which is the main focus of CFL.
\newpage

\begin{table}[th!]
\centering 
\tiny
\begin{minipage}[t]{0.48\textwidth}
\centering
\begin{tabular}{|p{2.8cm}|p{1.2cm}|p{1.5cm}|}
\hline
\textbf{Reference} & \textbf{Algorithm} & \textbf{Tiers} \\ \hline
Ghosh et al. 2019~\cite{ghosh2019robust} & N/A & A-D-G-I-N \\ \hline
Sattler et al. 2020~\cite{sattler2020clustered} & CFL / MTCFL & A-D-G-K-L-N \\ \hline
Mansour et al. 2020~\cite{mansour2020three} & HypCluster & A-E-G-J \\ \hline
Briggs et al. 2020~\cite{briggs2020federated} & FL+HC & B-D-G-I \\ \hline
Ghosh et al. 2020~\cite{IFCA} & IFCA & A-E-G-J \\ \hline
Dennis et al. 2021~\cite{dennis2021heterogeneity} & k-Fed & A-F-G-I-L \\ \hline
Li et al. 2021~\cite{flsc} & FLSC & A-E-H-J \\ \hline
Agrawal et al. 2021~\cite{geneticcfl} & GeneticCFL & A-F-G-J-M \\ \hline
Marfoq et al. 2021~\cite{FedEM} & FedEM & A-E-J \\ \hline
Duan et al. 2021~\cite{duan2021flexible} & FlexCFL & A-D-G-J-L \\ \hline
Jamali-Rad et al. 2022~\cite{FLT} & FLT & A-F-H-J \\ \hline
Gong et al. 2022~\cite{gong2022adaptive} & AutoCFL & A-E-G-J-M \\ \hline
Ruan et al. 2022~\cite{ruan2022fedsoft} & FedSoft & A-E-H-J \\ \hline
Long et al. 2023~\cite{long2023multi} & FeSEM & A-E-H-J-M \\ \hline
Jothimurugesan et al. 2023~\cite{FedDrift} & FedDrift & A-E-G-J \\ \hline
Jingyu et al. 2023~\cite{ascfl} & ASCFL & A-F-G-J \\ \hline
Yihan et al. 2023~\cite{ICFL} & ICFL & A-D-G-K \\ \hline
Tun et al. 2023~\cite{CP-CFL} & CP-CFL & C-E-G-J \\ \hline
\end{tabular}
\end{minipage}
\hfill
\begin{minipage}[t]{0.48\textwidth}
\centering
\begin{tabular}{|p{2.5cm}|p{1.2cm}|p{1.5cm}|}
\hline
\textbf{Reference} & \textbf{Algo.} & \textbf{Tiers} \\ \hline
Vahidian et al. 2023~\cite{PACFL} & PACFL & A-F-G-I-L \\ \hline
Cai et al. 2023~\cite{FedCE} & FedCE & A-E-H-J-M \\ \hline
Morafah et al. 2023~\cite{FLIS} & FLIS & C-D-G-J \\ \hline
Ma et al. 2023~\cite{FedCAM} & FedCAM & A-E-H-J-M \\ \hline
Liang et al. 2023~\cite{liang2023efficient} & FedEOC & A-D-G-I-M \\ \hline
Mehta et al. 2023~\cite{mehta2023greedy} & FLACC & A-D-G-J-M \\ \hline
Yu et al. 2024~\cite{FPFC} & FPFC & B-D-G-J-M \\ \hline
Zhang et al. 2024~\cite{MD-ICFL} & MD-ICFL & A-F-G-J \\ \hline
Gao et al. 2024~\cite{FedPEC} & FedPEC & A-F-G-J \\ \hline
Zhang et al. 2024~\cite{FedGK} & FedGK & C-D-G-K \\ \hline
Liu et al. 2024~\cite{CFLGT} & CFLGT & A-D-G-J \\ \hline
Guo et al. 2024~\cite{FedRC} & FedRC & A-E-H-J-M \\ \hline
Luo et al. 2024~\cite{ppcfl} & pp-CFL & A-F-H-J-L-N \\ \hline
Gong et al. 2024~\cite{HiCFL} & HiCFL & A-D-G-K \\ \hline
Ren et al. 2025~\cite{TDCFL} & TDCFL & A-F-G-J-L \\ \hline
Cheng et al. 2025~\cite{SnapCFL} & SnapCFL & C-D-G-I \\ \hline
Zeng et al. 2025~\cite{StoCFL} & StoCFL & A-D-G-K-M \\ \hline
Guo et al. 2025~\cite{HCFL+} & HCFL+ & A-E-H-J-K-M \\ \hline
\end{tabular}
\end{minipage}
\caption{CFL approaches with corresponding tier classification...} \label{tab:CFL_letters}
\end{table}
\normalsize

\subsection{Evaluating CFL methods}\label{sec:cfl_eval}
The evaluation of CFL algorithms requires a rigorous experimental framework comprising three key components: representative non-IID data scenarios, diverse benchmarks, and appropriate performance metrics. In this section, we analyze how the state-of-the-art addresses these challenges.

\subsubsection{Formalization of Non-IID Data in FL}\label{sec:statistical_heterogeneity}
Non-Independent and Identically Distributed (non-IID) data give rise to a broad range of statistical challenges. The taxonomy of non-IID heterogeneity, as proposed by Kairouz et al.~\citep{kairouz2021advances}, is often used as the foundation to form interpretable non-IID scenarios for evaluating CFL benchmarks. We therefore rely on the five categories of non-IID data proposed by this taxonomy. Before introducing it formally, we provide an intuitive overview of the five major forms of data heterogeneity in FL. These categories describe how clients' data differ from each other. First, clients may observe the same labels but with inputs that differ visually or structurally (e.g., rotations or style differences), which corresponds to Concept shift on feature. Second, clients may apply different labeling functions to identical inputs, resulting in Concept shift on targets. Clients can also differ in their feature statistics (e.g., brightness or texture, features distribution skew)  or in the frequency of classes they observe (label distribution skew). Finally, clients may simply have unequal dataset sizes, a form of heterogeneity known as quantity skew. These five types cover most non-IID situations encountered in practice and clarify how heterogeneity affects learning stability and cluster formation in CFL.

To formalize these situations, we consider a general supervised task where each client $i \in I = \{1,\ldots, N \}$ has a local dataset $D_i$ represented as a set of samples $D_i = \{(x^{(i)}_j, y^{(i)}_j) \mid 1 \leq j \leq n_i\}$ where $(x^{(i)}_{j})_{1 \leq j \leq n_i}$ represents the features and $(y^{(i)}_{j})_{1 \leq j \leq n_i}$ represents the target and with $|D_i| = n_i$. 

Using this notation, we define the random variables $X$ representing the possible outcome of features and $Y$ representing the possible outcome of the targets of all client datasets $(D_i)_{i\in I}$. That means that, in this case, for all $i\in I$, $(x^{(i)}_{j})_{1 \leq j \leq n_i}$ are realizations of $X$ and  $(y^{(i)}_{j})_{1 \leq j \leq n_i}$ are realizations of $Y$.

If we denote the probability space as $(\Omega, \mathcal{F}, P)$ where  $\Omega$ is the sample space, representing all possible outcomes of the random experiment associated with sampling from the dataset $D= \cup_{i=1}^{N}D_i$, $\mathcal{F}$ its associate sigma algebra and $\mathcal{P}$ the probability measure, then the joint random variable $(X,Y)$ can be defined as $(X,Y):\Omega \mapsto D = \cup_{i=1}^{N}D_i$. 

Each client's dataset $D_i$ is represented by its local joint distribution $\mathcal{P}_{i}(X, Y)$, which we can factor as follows using Bayes' rule: 
$\mathcal{P}_{i}(X, Y) = \mathcal{P}_{i}(Y) \mathcal{P}_{i}(X | Y) = \mathcal{P}_{i}(X) \mathcal{P}_{i}(Y | X)$ where $\mathcal{P}_{i}(X)$ is the marginal distribution of the features (respectively $\mathcal{P}_{i}(Y)$ the target) and $\mathcal{P}_{i}(Y | X)$ is the distribution of the target conditioned by the features (respectively $\mathcal{P}_{i}(X | Y)$ is the distribution of features conditioned by the target) for client $i \in I$. 

\begin{figure}[htbp]
    \adjustbox{width=\textwidth}{\includegraphics{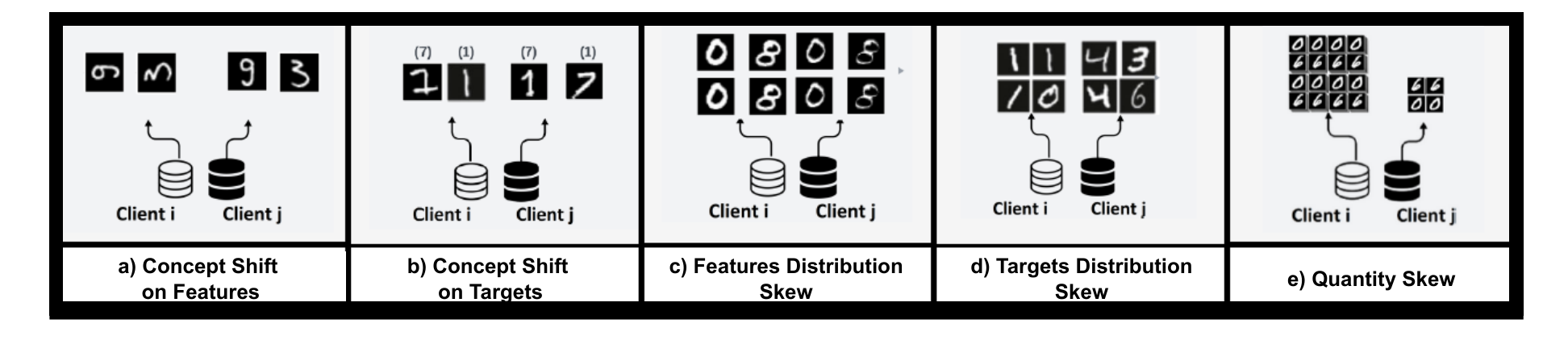}}
    \caption{Illustration of non-IID categories for two clients $i$ and $j$ using MNIST samples.}
    \label{fig:non_iid_categories}
\end{figure}

While this taxonomy was originally developed for classification problems, where the target variable $Y$ is discrete, it naturally extends to general supervised learning settings. However, interpreting these categories becomes less direct when $Y$ is continuous or structured. To make each situation concrete, we illustrate the five categories using MNIST (Figure~\ref{fig:non_iid_categories}), where different clients' image data exhibit distinct forms of heterogeneity. For two clients $i, j \in I$ with $i \neq j$, datasets $D_i$ and $D_j$ can be considered non-IID if one or more of the following conditions apply:

\begin{enumerate}[label=\alph*.]

\item \textbf{Concept shift\footnote{Also referred as ``Concept Drift''~\citep{kairouz2021advances, FedDrift, FedRC}} on features}:  
(same target, different features)  
This occurs when $\mathcal{P}_i(X|Y) \neq \mathcal{P}_j(X|Y)$ even though $\mathcal{P}_i(Y) = \mathcal{P}_j(Y)$.  
In other words, clients share the same target space data, still the feature representations differ.  
For example, in Figure~\ref{fig:non_iid_categories}(a), both clients share the same label distribution $\mathcal{P}(Y)$ (possessing both ``3''s and ``9''s). However, the conditional distribution $\mathcal{P}(X|Y)$ differs because Client $j$'s images are rotated relative to Client $i$'s, meaning the feature representation $X$ for the same label $Y$ has changed.

\item \textbf{Concept shift\footnotemark[3] on targets}\footnote{On labels in \citep{kairouz2021advances}}:  
(same features, different target)  
This occurs when $\mathcal{P}_i(Y|X) \neq \mathcal{P}_j(Y|X)$ even though $\mathcal{P}_i(X) = \mathcal{P}_j(X)$. 
This means that, even when clients share the same feature inputs, the target mapping is different. 
For instance, in Figure~\ref{fig:non_iid_categories}(b), both clients observe identical feature inputs $X$ (e.g., the image of a ``7''). However, their mapping $\mathcal{P}(Y|X)$ diverges: Client $i$ correctly labels it as ``7'', while Client $j$ labels the same feature as ``1'' (label swapping).

\item \textbf{Feature distribution skew}:  
This occurs when $\mathcal{P}_i(X) \neq \mathcal{P}_j(X)$ even if $\mathcal{P}_i(Y|X) = \mathcal{P}_j(Y|X)$.
In this case, clients share the same underlying concepts, but the overall distribution of feature values is skewed.
For instance, in Figure~\ref{fig:non_iid_categories}(c), both clients agree on the concept (Both ``0'' and ``8'' look like the actual digits). However, the marginal feature distribution $\mathcal{P}(X)$ differs: Client $i$ exclusively holds images drawn with a thick stroke, while Client $j$ holds images with a thin stroke, representing a skew in the feature domain (e.g., writing style).

\item \textbf{Target\footnotemark[4] distribution skew}:  
$\mathcal{P}_i(Y) \neq \mathcal{P}_j(Y)$ even though $\mathcal{P}_i(X|Y) = \mathcal{P}_j(X|Y)$.  
Here, clients share similar feature-target relationships but differ in the frequencies of target values.  
For instance, in Figure~\ref{fig:non_iid_categories}(d), the marginal label distribution $\mathcal{P}(Y)$ varies significantly: Client $i$'s dataset is dominated by the label ``1'', whereas Client $j$ possesses almost exclusively ``4''s.

\item \textbf{Quantity skew}:  
The number of samples varies significantly among clients, such that $|D_i| \ll |D_j|$ or vice versa.  
This imbalance, illustrated in Figure~\ref{fig:non_iid_categories}(e), causes clients to contribute unequally to the global model update, potentially biasing the aggregation towards clients with larger local datasets $|D_j|$.

\end{enumerate}

This taxonomy offers a structured framework to characterize how client data diverges. Since real-world FL data are typically private and inherently non-IID, early CFL algorithms like IFCA~\citep{IFCA} already relied on this taxonomy to synthetically generate heterogeneous scenarios from centralized datasets. By partitioning and transforming data, such as applying rotations to image datasets like MNIST or CIFAR, researchers can simulate client groups exhibiting distinct yet interpretable forms of heterogeneity. For instance, assigning rotations of 0°, 90°, 180°, and 270° to different client groups creates a synthetic benchmark for evaluating CFL algorithms in concept shift on features scenarios.

\begin{figure}[htbp]
\centering
\begin{subfigure}[b]{0.48\textwidth}
\centering
\includegraphics[width=0.7\linewidth]{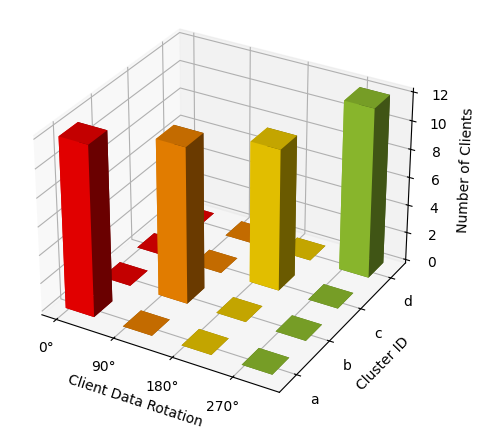}
\caption{Optimal clients-to-clusters assignment, groups reflecting underlying data heterogeneity.}
\label{fig:optimal_assignement}
\end{subfigure}
\hfill
\begin{subfigure}[b]{0.48\textwidth}
\centering
\includegraphics[width=0.7\linewidth]{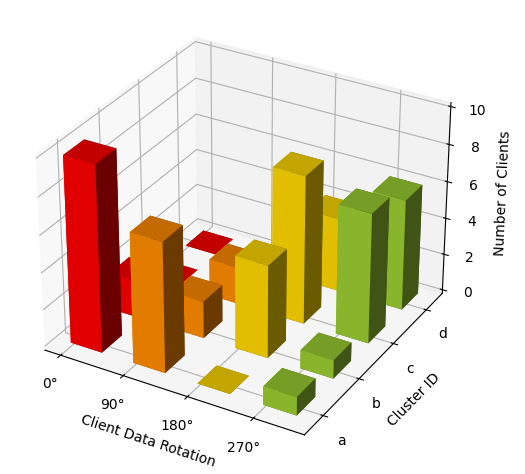}
\caption{Poor clients-to-clusters assignment, failing to reflect underlying heterogeneity.}
\label{fig:poor_assignement}
\end{subfigure}
\caption{Illustration of client grouping in CFL under rotation-induced data heterogeneity: optimal vs. poor clustering}
\label{fig:metrics_illustration}
\end{figure}

In these settings, traditional FL methods often produce biased models that fail to generalize, whereas CFL algorithms can effectively cluster clients with similar data distributions without direct access to raw samples. An optimal CFL approach should thus separate these distributions into coherent clusters, as illustrated in Figure~\ref{fig:metrics_illustration}.
To make this clearer, the figure contrasts two clustering outcomes under rotation-induced heterogeneity. In the left panel, clients sharing the same rotation are grouped consistently, leading to well-separated clusters that reflect the true underlying data distributions. In the right panel, however, clients with the same rotation are scattered across different clusters, illustrating a poor assignment that fails to capture the structure of the data. This mismatch exemplifies the core challenge in CFL: ensuring that clients with similar distributions are assigned to the same cluster. 

Nevertheless, several studies highlight nuanced challenges. Mehta et al.~\cite{mehta2023greedy}  observed that ambiguous cluster boundaries may emerge when feature distribution skew dominates, making cluster separation unclear. Similarly, Guo et al.~\cite{FedRC} noted that clustered FL methods sometimes group clients primarily by target(label) similarity, which can lead to overfitting local distributions, particularly in the presence of target(label) distribution skew combined with other heterogeneities.

Even though these concepts are useful for classification, extending them beyond discrete targets introduces ambiguity. In tasks such as regression, forecasting, or other supervised problems, the target space is often continuous or multidimensional. As a result, differences in $\mathcal{P}_i(Y)$ or $\mathcal{P}_i(Y|X)$ cannot be categorized discretely.

Developing non-IID benchmarks for such domains may require
domain-specific formulations or relies on natural heterogeneity, as seen in distributed IoT setups (e.g. geographical influence or similar usage patterns). In these cases, evaluating CFL becomes more complex, and benchmarks tend to focus mainly on improving performance metrics. Identifying optimal clusters is challenging, even in controlled benchmarks where CFL algorithms are tested with available ground-truth data.

Nevertheless, this taxonomy remains a practical foundation for understanding statistical heterogeneity in CFL. It provides a consistent basis for designing controlled and reproducible experimental setups across various tasks.

\subsubsection{Datasets and Tasks}\label{sec:datasets_and_problems}

The CFL solutions reviewed in Section~\ref{sec:clf_classification} all provide experimental evaluations to benchmark their approaches. As mentioned in Section~\ref{sec:statistical_heterogeneity}, the data heterogeneity taxonomy plays an important role in assessing CFL performance. In this section, we examine the datasets, tasks, and types of heterogeneity used to benchmark these solutions. Specifically, we analyze which datasets have been employed and which heterogeneity categories the proposed methods were evaluated on. This categorization highlights gaps in the literature and points toward potential directions for future research.

Although the non-IID taxonomy is commonly used to transform and distribute centralized datasets into federated settings, many studies do not explicitly map their experimental designs to this taxonomy. Nevertheless, in most cases, their configurations align well with it. Some papers lack a unified terminology to describe similar concepts. For instance, \cite{mehta2023greedy} uses the terms concept drift and label distribution skew, while \cite{FedRC} refers to concept shift and label distribution shift, respectively. Such discrepancies may introduce ambiguity for the reader.

To reduce this ambiguity and establish a consistent basis for comparison, Table~\ref{tab:datasets-noniid} summarizes the different benchmarks found in the literature, aligns them with the non-IID taxonomy defined in Section~\ref{sec:statistical_heterogeneity}. Additionally, it explains how heterogeneity was created for each dataset. We believe that providing a unified framework across studies facilitates clearer comparisons.

In Table~\ref{tab:datasets-noniid}, the column \textit{"Taxonomy"}shows each non-iid scenario from Kairouz et al.\citep{kairouz2021advances}.  The \textit{"Task" }column specifies the learning task addressed (e.g., classification, prediction, regression). "Datasets" lists the public datasets used (e.g., MNIST, FEMNIST, Shakespeare). The \textit{"Heterogeneity creation"} column describes how data heterogeneity is introduced into datasets, while the \textit{"Papers"} column enumerates the corresponding algorithm/publications.

\tiny
\setlength{\tabcolsep}{3pt}
\begin{longtable}{|p{0.5in}|p{0.5in}|p{1in}|p{1.2in}|p{2in}|}
\caption{Taxonomy-based view of Tasks, Datasets, and Heterogeneity creation methods in CFL literature.} \label{tab:datasets-noniid}
\\
\hline
\textbf{Taxonomy} & \textbf{Task} & \textbf{Datasets} & \textbf{Heterogeneity Creation} & \textbf{Papers} \\ \hline
\endfirsthead

\hline
\textbf{Taxonomy} & \textbf{Task} & \textbf{Datasets} & \textbf{Heterogeneity Creation} & \textbf{Papers} \\ \hline
\endhead

\hline
\multicolumn{5}{r}{\textit{Continued on next page}} \\
\endfoot

\hline
\endlastfoot

\multirow[t]{10}{=}{\textbf{Concept Shift on Features}} 
& Image Class. 
& MNIST, EMNIST, F-MNIST, CIFAR-10 
& Image Rotations 
& IFCA\cite{IFCA}, k-FED \cite{dennis2021heterogeneity}, AutoCFL \cite{gong2022adaptive}, ASCFL \cite{ascfl}, ICFL \cite{ICFL}, CP-CFL \cite{CP-CFL}, FedCE \cite{FedCE}, FLACC \cite{mehta2023greedy}, MD-ICFL \cite{MD-ICFL}, HiCFL \cite{HiCFL}, SnapCFL \cite{SnapCFL}, StoCFL \cite{StoCFL} \\ \cline{2-5}

& Image Class. 
& MNIST, USPS, SVHN, SIGN 
& Separate Datasets per client 
& PACFL \cite{PACFL}, MD-ICFL \cite{MD-ICFL} \\ \cline{2-5}

& Regression 
& Housing, Bodyfat 
& Separate Datasets per client 
& FPFC \cite{FPFC} \\ \hline

\multirow[t]{8}{=}{\textbf{Concept Shift on Targets}} 
& Image Class. 
& MNIST, F-MNIST, CIFAR-10, CIFAR-100, Tiny-ImageNet, FEMNIST 
& Label Swapping 
& MTCFL \cite{sattler2020clustered}, FPFC \cite{FPFC}, StoCFL \cite{StoCFL}, FL+HC \cite{briggs2020federated}, FedGK \cite{FedGK}, FedRC \cite{FedRC}, HCFL+ \cite{HCFL+}, FLACC \cite{mehta2023greedy}, MD-ICFL \cite{MD-ICFL} \\ \cline{2-5}

& Rating Pred. 
& ml-100K, ml-1M 
& Split by User (Rating bias) 
& FEDEOC \cite{liang2023efficient} \\ \hline

\multirow[t]{14}{=}{\textbf{Feature Distribution Skew}} 
& Image Class. 
& EMNIST, FEMNIST, FedCelebA 
& Natural Split (Users/Writers) 
& HypCluster \cite{mansour2020three}, k-FED \cite{dennis2021heterogeneity}, FL+HC \cite{briggs2020federated}, IFCA \cite{IFCA}, FlexCFL \cite{duan2021flexible}, FedEM \cite{FedEM}, FeSEM \cite{long2023multi}, FLACC \cite{mehta2023greedy}, FedPEC \cite{FedPEC}, StoCFL \cite{StoCFL} \\ \cline{2-5}

& Image Class. 
& EMNIST 
& Uppercase vs. Lowercase 
& FedSoft \cite{ruan2022fedsoft}, FedCE \cite{FedCE}, FLACC \cite{mehta2023greedy} \\ \cline{2-5}

& Image Class. 
& F-MNIST, CIFAR-10, CIFAR-100, Tiny-ImageNet 
& Image Corruption 
& FedRC \cite{FedRC}, HCFL+ \cite{HCFL+} \\ \cline{2-5}

& NLP / HAR / Sent. Analysis 
& Shakespeare, AG News, USC-HAD, Sentiment140 
& Natural Split (Actors/Topic/Users) 
& k-FED \cite{dennis2021heterogeneity}, FedEM \cite{FedEM}, MTCFL \cite{sattler2020clustered}, HiCFL \cite{HiCFL}, FedPEC \cite{FedPEC} \\ \hline

\textbf{Target Distribution Skew} 
& Image Class. 
& MNIST, EMNIST, FEMNIST, F-MNIST, CIFAR-10, CIFAR-100, STL-10, SVHN, Tiny-ImageNet 
& Different Label Distribution for each client
& FLSC \cite{flsc}, FlexCFL \cite{duan2021flexible}, GeneticCFL \cite{geneticcfl}, AutoCFL \cite{gong2022adaptive}, FLT \cite{FLT}, FPFC \cite{FPFC}, MD-ICFL \cite{MD-ICFL}, pp-CFL \cite{ppcfl}, HiCFL \cite{HiCFL}, SnapCFL \cite{SnapCFL}, FedEM \cite{FedEM}, k-FED \cite{dennis2021heterogeneity}, PACFL \cite{PACFL}, FLIS \cite{FLIS}, FedCAM \cite{FedCAM}, FedGK \cite{FedGK}, CFLGT \cite{CFLGT}, FedRC \cite{FedRC}, HCFL+ \cite{HCFL+}, TDCFL \cite{TDCFL}, CP-CFL \cite{CP-CFL} \\ \hline

\textbf{Quantity Skew} 
& Image Class. 
& MNIST, EMNIST, FEMNIST, F-MNIST, CIFAR-10 
& Random Sample Size 
& AutoCFL \cite{gong2022adaptive}, FLACC \cite{mehta2023greedy} \\ \hline

\end{longtable}
\normalsize
Observations from Table~\ref{tab:datasets-noniid} provide insights into how different forms of heterogeneity are generated. Concept shift on features is typically induced by rotating images across clients for image classification tasks or by assigning different but related datasets to different clients (e.g., the Housing and Body Fat datasets). Concept shift on targets often arises from label swapping in classification problems, except for rating prediction tasks such as ml-100K and ml-1M~, where each user naturally exhibits unique preferences. Feature distribution skew is generally inherent to the dataset, stemming from user- or topic-specific distributions. The only artificial case is the use of image corruption by FedRC~\cite{FedRC} and HCFL+~\cite{HCFL+}. Conversely, target distribution skew is introduced by assigning clients different label distributions within classification tasks. Finally, quantity skew appears in a few works, where clients receive varying numbers of samples without further heterogeneity.
\begin{figure}[htb!]
    \centering
     \begin{subfigure}[b]{0.50\textwidth}
        \centering
        \includegraphics[width=\textwidth]{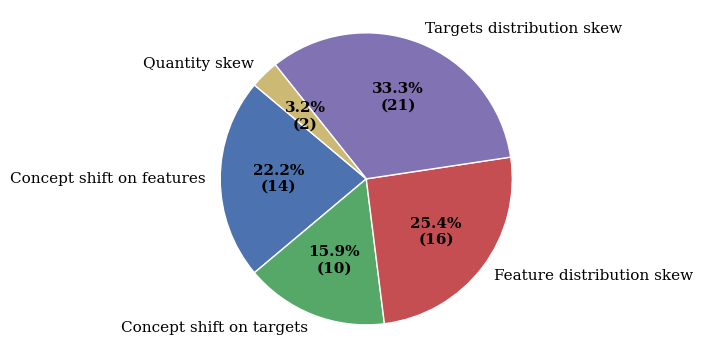}
        \caption{Distribution of the taxonomy Benchmarked Datasets}
        \label{fig:pie_taxonomy}
    \end{subfigure}
    \hfill
    \begin{subfigure}[b]{0.4\textwidth}
        \centering
        \includegraphics[width=0.8\textwidth]{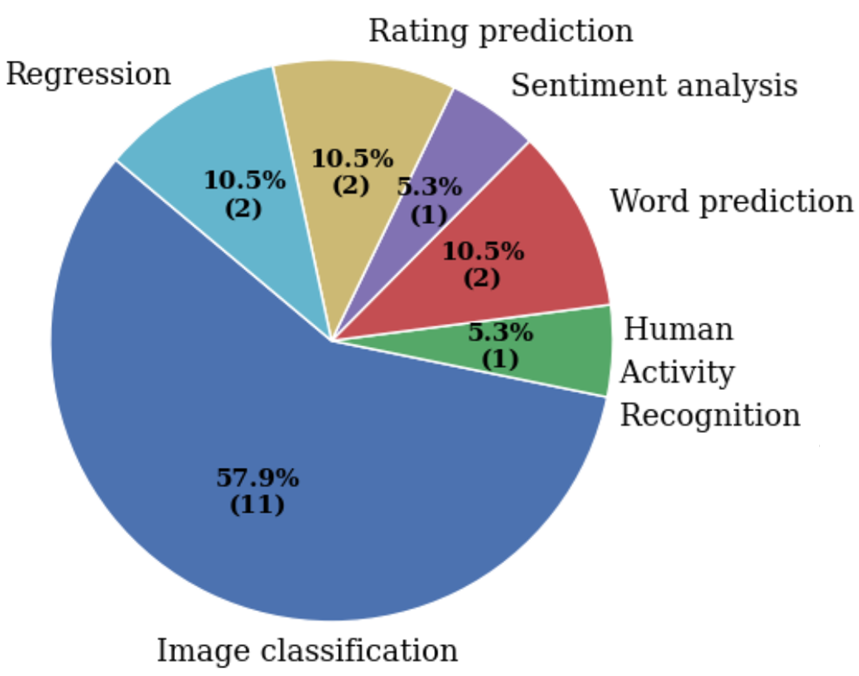}
        \caption{Distribution of Tasks of Benchmarked Datasets}
        \label{fig:pie_task}
    \end{subfigure}
    \caption{Distribution of experimental designs across reviewed core CFL papers from Table~\ref{tab:CFL_classification}.}
    \label{fig:piecharts}
\end{figure}

Figure~\ref{fig:pie_taxonomy} illustrates the distribution of heterogeneity types (taxonomy) explored across surveyed CFL algorithms. Each category—such as feature distribution skew, target distribution skew, or concept shift—is represented proportionally to the number of algorithms addressing it. The visualization shows that most studies focus on target or feature distribution skew, while quantity skew remains largely unexplored, having been evaluated only partially in AutoCFL~\cite{gong2022adaptive} and FLACC~\cite{mehta2023greedy}.

Figure~\ref{fig:pie_task} shows the distribution of learning task, emphasizing that image classification dominates as the primary benchmark. Only a few works consider regression, recommendation, or text-based prediction tasks. This imbalance reflects the convenience of applying heterogeneity taxonomies to image classification and highlights the need for broader empirical validation across task types. This trend largely stems from the ease of introducing heterogeneity in image data, such as rotations (concept shift on features) or corruptions (feature distribution skew), as shown for MNIST and CIFAR variants. In contrast, text, time-series, and tabular datasets appear less frequently, indicating a limited exploration of non-vision domains. 

\subsubsection{Performance and Clustering Metrics}\label{sec:metrics}
It is interesting to note that the majority of CFL algorithms rely primarily on task-specific metrics—such as average accuracy or RMSE across clients. Additionally, classical convergence indicators like average loss evolution curves are often used to demonstrate communication improvements over baselines. Very few studies introduce metrics specific to the clustering process itself. 

Some approaches, such as FeSEM~\cite{long2023multi}, CP-CFL~\cite{CP-CFL}, and FedCAM~\cite{FedCAM}, distinguish between \textit{micro} (weighted average based on data amount) and \textit{macro} (simple average across devices) aggregation of these results, though this distinction applies to FL in general rather than CFL specifically.

Regarding clustering-specific behaviors, a limited number of works analyze specific CFL criteria. For instance, FlexCFL~\cite{duan2021flexible} tracks the \textit{average discrepancy} (the distance between local and global cluster models), demonstrating that successful clustering lowers this metric over rounds. Similarly, AutoCFL~\cite{gong2022adaptive} analyzes the variance metrics across clients to validate the reduced heterogeneity within formed clusters.

However, explicit evaluation of cluster structural quality (against a ground truth) remains rare, likely because clustering in CFL is often treated as a heuristic to improve target metrics rather than an end goal. Notable exceptions include ICFL~\cite{ICFL}, FPFC~\cite{FPFC}, MD-ICFL~\cite{MD-ICFL}, and FedPEC~\cite{FedPEC}, which report the \textbf{Adjusted Rand Index (ARI)} or \textbf{Cluster Purity}, and TDCFL~\cite{TDCFL} which utilizes the \textbf{Silhouette Score}. Importantly, these studies highlight a strong correlation between structural metrics and model performance; for example, Yan et al.~\cite{ICFL} observes that algorithms achieving a perfect Purity (perfect recovery of ground-truth partitions) consistently yield the best performances.

\subsubsection{Timeline of CFL experimental comparison}

\begin{figure}[t!]
    \centering
    \tiny
\begin{subfigure}[b]{\textwidth}
        \centering
        
    \begin{tikzpicture}[scale=0.6,
        timeline/.style={thick,->,>=stealth},
        event/.style={rectangle, draw, fill=gray!20, rounded corners, minimum height=1em, text centered,  font=\fontsize{4.5}{4.5}\selectfont,  yshift=-10pt}]
        
        \draw[timeline] (-5,0) -- (17,0);
        \foreach \x/\y in {-5/2019, -4/2020, 0/2021, 4/2022, 8/2023, 12/2024, 16/2025} {
            \draw (\x,0.2) -- (\x,-0.2) node[below] {\y};
        }
            \node[event, fill=red!20] at (-5,1) {MTCFL~\cite{sattler2020clustered}};

            \node[event, fill = blue!20] at (0,1) {FedEM \cite{FedEM}};
            \node[event, fill = blue!20] at (0,1.7) {FeSEM \cite{long2023multi}};
            \node[event] at (4,1) {FedDrift \cite{FedDrift}};
            \node[event] at (4,1.7) {FLT \cite{FLT}};

            \node[event] at (7.5,1) {ASCFL \cite{ascfl}};
            \node[event] at (7.5,1.7) {ICFL \cite{ICFL}};
            \node[event] at (7.5,2.4) {FLIS \cite{FLIS}};
            \node[event] at (9.1,1) {FLACC \cite{mehta2023greedy}};
            \node[event] at (9,1.7) {AutoCFL \cite{gong2022adaptive}};
            \node[event] at (12,1) {FPFC \cite{FPFC}};
            \node[event] at (12,1.7) {MD-ICFL \cite{MD-ICFL}};
            \node[event] at (12,2.4) {FedRC \cite{FedRC}};
            
            \node[event] at (16,1) {HCFL+ \cite{HCFL+}};
            \node[event] at (16,1.7) {TDCFL \cite{TDCFL}};
            \node[event] at (16,2.4) {SnapCFL \cite{SnapCFL}};
            \node[event,fill = blue!20] at (16,3.1) {StoCFL \cite{StoCFL}};
            
        \end{tikzpicture}
    \end{subfigure}

    \begin{subfigure}[b]{\textwidth}
        \centering
    \begin{tikzpicture}[scale=0.6,
        timeline/.style={thick,->,>=stealth},
        event/.style={rectangle, draw, fill=gray!20, rounded corners, minimum height=1em, text centered,  font=\fontsize{4.5}{4.5}\selectfont, yshift=-10pt}]
        
        \draw[timeline] (-5,0) -- (17,0);
        \foreach \x/\y in {-5/2019, -4/2020, 0/2021, 4/2022, 8/2023, 12/2024, 16/2025} {
            \draw (\x,0.2) -- (\x,-0.2) node[below] {\y};
        }
            \node[event, fill=red!20] at (-4,1) {IFCA~\cite{IFCA}};

            \node[event] at (0,1) {FLSC~\cite{flsc}};
            \node[event] at (0,1.7) {GeneticCFL \cite{geneticcfl}};            
            \node[event] at (0,2.4) {k-FED~\cite{dennis2021heterogeneity}};
            \node[event,fill = blue!20] at (0,3.1) {FlexCFL \cite{duan2021flexible}};

            \node[event] at (4,1) {FedDrift \cite{FedDrift}};
            \node[event] at (4,1.7) {FLT \cite{FLT}};
            \node[event,fill = blue!20] at (4,2.4) {FedSoft \cite{ruan2022fedsoft}};
            \node[event] at (4,3.1) {AutoCFL \cite{gong2022adaptive}};
            
            \node[event,fill = blue!20] at (7.3,1) {ICFL \cite{ICFL}};
            \node[event] at (7.3,1.7) {FLIS \cite{FLIS}};
            \node[event] at (7.3,2.4) {CP-CFL \cite{CP-CFL}};
            \node[event,fill = blue!20] at (7.3,3.1) {PACFL \cite{PACFL}};
            \node[event] at (9.1,1.7) {FedCAM \cite{FedCAM}};
            \node[event] at (9.1,2.4) {FedEOC \cite{liang2023efficient}};
            \node[event] at (9.1,1) {FLACC \cite{mehta2023greedy}};
            
            \node[event] at (11.5,1.7) {FedPEC \cite{FedPEC}};
            \node[event] at (11.5,1) {FPFC \cite{FPFC}};
            \node[event] at (11.5,2.4) {MD-ICFL \cite{MD-ICFL}};
            \node[event] at (13.1,1) {CFLGT \cite{CFLGT}};
            \node[event] at (13.1,1.7) {pp-CFL \cite{ppcfl}};
            \node[event,fill = blue!20] at (13.15,2.4) {FedRC \cite{FedRC}};

            \node[event] at (16,1) {SnapCFL \cite{SnapCFL}};
            \node[event] at (16,1.7) {HCFL+ \cite{HCFL+}};
            \node[event] at (16,2.4) {TDCFL \cite{TDCFL}};
            \node[event,fill = blue!20] at (16,3.1) {StoCFL \cite{StoCFL}};

        \end{tikzpicture}
    \end{subfigure}
    
    \begin{subfigure}[b]{\textwidth}
    \centering
    \begin{tikzpicture}[scale=0.6,
        timeline/.style={thick,->,>=stealth},
        event/.style={rectangle, draw, fill=gray!20, rounded corners, minimum height=1em, text centered,  font=\fontsize{4.5}{4.5}\selectfont, yshift=-10pt}]
        
        \draw[timeline] (-5,0) -- (17,0);
        \foreach \x/\y in {-5/2019, -4/2020, 0/2021, 4/2022, 8/2023, 12/2024, 16/2025} {
            \draw (\x,0.2) -- (\x,-0.2) node[below] {\y};
        }
        \node[event, fill=red!20] at (-4,1) {HypCluster~\cite{mansour2020three}};

        \node[event, fill=blue!20] at (8,1) {ICFL \cite{ICFL}};
        \node[event] at (4,1) {FeSEM \cite{long2023multi}};
    \end{tikzpicture}
\end{subfigure}
 \begin{subfigure}[b]{\textwidth}
        \centering
    \begin{tikzpicture}[scale=0.6,
        timeline/.style={thick,->,>=stealth},
        event/.style={rectangle, draw, fill=gray!20, rounded corners, minimum height=1em, text centered,  font=\fontsize{4.5}{4.5}\selectfont, yshift=-10pt}]
        
        \draw[timeline] (-5,0) -- (17,0);
        \foreach \x/\y in {-5/2019, -4/2020, 0/2021, 4/2022, 8/2023, 12/2024, 16/2025} {
            \draw (\x,0.2) -- (\x,-0.2) node[below] {\y};
        }
            \node[event, fill=red!20] at (-4,1) {FL+HC~\cite{briggs2020federated}};

            \node[event] at (4,1) {AutoCFL~\cite{gong2022adaptive}};
            \node[event] at (8,1) {ASCFL \cite{ascfl}};
            \node[event] at (12,1.7) {FLACC \cite{mehta2023greedy}};
            \node[event] at (12,1) {CFLGT \cite{CFLGT}};
        \end{tikzpicture}
    \end{subfigure}

\begin{subfigure}[b]{\textwidth}
    \centering
    \begin{tikzpicture}[scale=0.6,
        timeline/.style={thick,->,>=stealth},
        event/.style={rectangle, draw, fill=gray!20, rounded corners, minimum height=1em, text centered,  font=\fontsize{4.5}{4.5}\selectfont, yshift=-10pt}]
        
        \draw[timeline] (-5,0) -- (17,0);
        \foreach \x/\y in {-5/2019, -4/2020, 0/2021, 4/2022, 8/2023, 12/2024, 16/2025} {
            \draw (\x,0.2) -- (\x,-0.2) node[below] {\y};
        }
        \node[event, fill=red!20] at (0,1) {FedEM \cite{FedEM}};
        \node[event] at (4,1) {FedSoft \cite{ruan2022fedsoft}};
        \node[event] at (12,1) {FedRC \cite{FedRC}};
    \end{tikzpicture}
\end{subfigure}

\begin{subfigure}[b]{\textwidth}
        \centering
    \begin{tikzpicture}[scale=0.6,
        timeline/.style={thick,->,>=stealth},
        event/.style={rectangle, draw, fill=gray!20, rounded corners, minimum height=1em, text centered,  font=\fontsize{4.5}{4.5}\selectfont, yshift=-10pt}]
        
        \draw[timeline] (-5,0) -- (17,0);
        \foreach \x/\y in {-5/2019, -4/2020, 0/2021, 4/2022, 8/2023, 12/2024, 16/2025} {
            \draw (\x,0.2) -- (\x,-0.2) node[below] {\y};
        }
            \node[event, fill=red!20] at (-0.5,1) {FeSEM \cite{long2023multi}};

            \node[event, fill=blue!20] at (1.1,1) {FlexCFL \cite{duan2021flexible}};
            \node[event] at (4,1) {FLT \cite{FLT}};
            \node[event] at (8,1) {FedCAM \cite{FedCAM}};
            \node[event, fill=blue!20] at (8,1.7) {ICFL \cite{ICFL}};
            \node[event] at (12,1) {CFLGT \cite{CFLGT}};
            \node[event, fill=blue!20] at (12,1.7) {FedRC \cite{FedRC}};
            \node[event] at (16,1) {HCFL+ \cite{HCFL+}};
            \node[event, fill=blue!20] at (16,1.7) {StoCFL \cite{StoCFL}};
            
        \end{tikzpicture}
    \end{subfigure}
     
\begin{subfigure}[b]{\textwidth}
    \centering
    \begin{tikzpicture}[scale=0.6,
        timeline/.style={thick,->,>=stealth},
        event/.style={rectangle, draw, fill=gray!20, rounded corners, minimum height=1em, text centered,  font=\fontsize{4.5}{4.5}\selectfont, yshift=-10pt}]
        
        \draw[timeline] (-5,0) -- (17,0);
        \foreach \x/\y in {-5/2019, -4/2020, 0/2021, 4/2022, 8/2023, 12/2024, 16/2025} {
            \draw (\x,0.2) -- (\x,-0.2) node[below] {\y};
        }
        \node[event, fill=red!20] at (0,1) {FlexCFL \cite{duan2021flexible}};
        \node[event] at (12,1) {FedGK \cite{FedGK}};
        \node[event] at (16,1) {TDCFL \cite{TDCFL}};
    \end{tikzpicture}
\end{subfigure}

\begin{subfigure}[b]{\textwidth}
    \centering
    \begin{tikzpicture}[scale=0.6,
        timeline/.style={thick,->,>=stealth},
        event/.style={rectangle, draw, fill=gray!20, rounded corners, minimum height=1em, text centered,  font=\fontsize{4.5}{4.5}\selectfont, yshift=-10pt}]
        
        \draw[timeline] (-5,0) -- (17,0);
        \foreach \x/\y in {-5/2019, -4/2020, 0/2021, 4/2022, 8/2023, 12/2024, 16/2025} {
            \draw (\x,0.2) -- (\x,-0.2) node[below] {\y};
        }
        \node[event, fill=red!20] at (4,1) {FedSoft \cite{ruan2022fedsoft}};
        \node[event] at (8,1) {FedCE \cite{FedCE}};
    \end{tikzpicture}
\end{subfigure}

\begin{subfigure}[b]{\textwidth}
\centering
    \begin{tikzpicture}[scale=0.6,
        timeline/.style={thick,->,>=stealth},
        event/.style={rectangle, draw, fill=gray!20, rounded corners, minimum height=1em, text centered,  font=\fontsize{4.5}{4.5}\selectfont, yshift=-10pt}]
        
        \draw[timeline] (-5,0) -- (17,0);
        \foreach \x/\y in {-5/2019, -4/2020, 0/2021, 4/2022, 8/2023, 12/2024, 16/2025} {
            \draw (\x,0.2) -- (\x,-0.2) node[below] {\y};
        }
        \node[event, fill=red!20] at (8,1) {PACFL \cite{PACFL}};
        \node[event] at (12,1) {FPFC \cite{FPFC}};
    \end{tikzpicture}
\end{subfigure}

\begin{subfigure}[b]{\textwidth}
\centering
    \begin{tikzpicture}[scale=0.6,
        timeline/.style={thick,->,>=stealth},
        event/.style={rectangle, draw, fill=gray!20, rounded corners, minimum height=1em, text centered,  font=\fontsize{4.5}{4.5}\selectfont, yshift=-10pt}]

        \draw[timeline] (-5,0) -- (17,0);
        \foreach \x/\y in {-5/2019, -4/2020, 0/2021, 4/2022, 8/2023, 12/2024, 16/2025} {
            \draw (\x,0.2) -- (\x,-0.2) node[below] {\y};
        }
        \node[event, fill=red!20] at (8,1) {ICFL \cite{ICFL}};

        \node[event] at (16,1) {HCFL+ \cite{HCFL+}};
    \end{tikzpicture}
\end{subfigure}

\begin{subfigure}[b]{\textwidth}
    \centering
    \begin{tikzpicture}[scale=0.6,
        timeline/.style={thick,->,>=stealth},
        event/.style={rectangle, draw, fill=gray!20, rounded corners, minimum height=1em, text centered,  font=\fontsize{4.5}{4.5}\selectfont, yshift=-10pt}]
        
        \draw[timeline] (-5,0) -- (17,0);
        \foreach \x/\y in {-5/2019, -4/2020, 0/2021, 4/2022, 8/2023, 12/2024, 16/2025} {
            \draw (\x,0.2) -- (\x,-0.2) node[below] {\y};
        }
        \node[event, fill=red!20] at (15.5,1) {StoCFL \cite{StoCFL}};

        \node[event] at (17.1,1) {HCFL+ \cite{HCFL+}};
        
    \end{tikzpicture}
\end{subfigure}
\begin{subfigure}[b]{\textwidth}
    \centering
\begin{tikzpicture}[scale=0.6,
        timeline/.style={thick,->,>=stealth},
        event/.style={rectangle, draw, fill=gray!20, rounded corners, minimum height=1em, text centered,  font=\fontsize{4.5}{4.5}\selectfont, yshift=-10pt}]
        
        \draw[timeline] (-5,0) -- (17,0);
        \foreach \x/\y in {-5/2019, -4/2020, 0/2021, 4/2022, 8/2023, 12/2024, 16/2025} {
            \draw (\x,0.2) -- (\x,-0.2) node[below] {\y};
        }
        \node[event, fill=red!20] at (15.5,1) {FedRC \cite{FedRC}};

        \node[event] at (17.1,1) {HCFL+ \cite{HCFL+}};
        
    \end{tikzpicture}
\end{subfigure}
    \caption{Timelines of Experimental Comparisons Between CFL Algorithms. Each timeline highlights a baseline algorithm (shown in red) and the subsequent challenger's methods that claim superior performance over it under limited experimental conditions. Algorithms that serve as baselines in their own respective timelines are shown in blue when they appear in previous timelines. This visualization illustrates how CFL methods evolve through targeted, often narrow, comparisons rather than comprehensive benchmarking across the field.}
    \label{fig:cfl_timeline}
    \end{figure}

To better understand the evolution and benchmarking practices in CFL, we summarize in Figure~\ref{fig:cfl_timeline} the timeline of experimental comparisons of algorithms listed in Table~\ref{tab:CFL_classification}. Each timeline focuses on a key algorithm (highlighted in red) that has served as a reference point for subsequent works. As shown in the figure, both MTCFL (challenged by 16 follow-up studies) and IFCA (respectively by 25) serve as benchmarks for the majority of algorithms, underlining their major influence and pioneering roles in the development of server-side and client-side CFL strategies. It is noteworthy that no metadata-based methods are challenged in these timelines, likely because such approaches are highly context-dependent, less domain-agnostic, and may raise additional privacy concerns. Among foundational works, HypCluster~\cite{mansour2020three} has only two direct competitors—an understandable outcome given its conceptual similarity to IFCA and its status as a non-peer-reviewed paper—while FL+HC~\cite{briggs2020federated} is compared against only four methods. 

Among later contributions, FeSEM~\cite{long2023multi} is the only subsequent method that distinguishes itself, being confronted by eight algorithms. As an early pioneer of soft clustering and personalized CFL, it demonstrated the value of integrating both mechanisms into CFL frameworks. Unfortunately, after that, no algorithm clearly stands out as a reference to be surpassed by subsequent CFL approaches. This illustrates one of the field’s main current challenges: despite incremental progress, as CFL research has matured, the proliferation of specialized algorithms targeting narrow contexts has made it increasingly difficult to identify clear leaders or universally effective methods. Another limitation also emerges when examining experimental comparisons. While many studies benchmark CFL algorithms against previous baselines, the distinction between server-side, client-side, or metadata-based approaches is rarely made explicit. As a result, CFL methods often compare performance metrics such as accuracy or convergence speed without addressing the cost asymmetry between these paradigms. Consequently, the abundance of state-of-the-art solutions built on distinct assumptions and scenarios complicates the selection of an appropriate CFL algorithm for both researchers and practitioners.

\subsection{Review of CFL Applications}
\label{sec:app}

\begin{figure}[ht!]
    \centering
    \includegraphics[width=\linewidth]{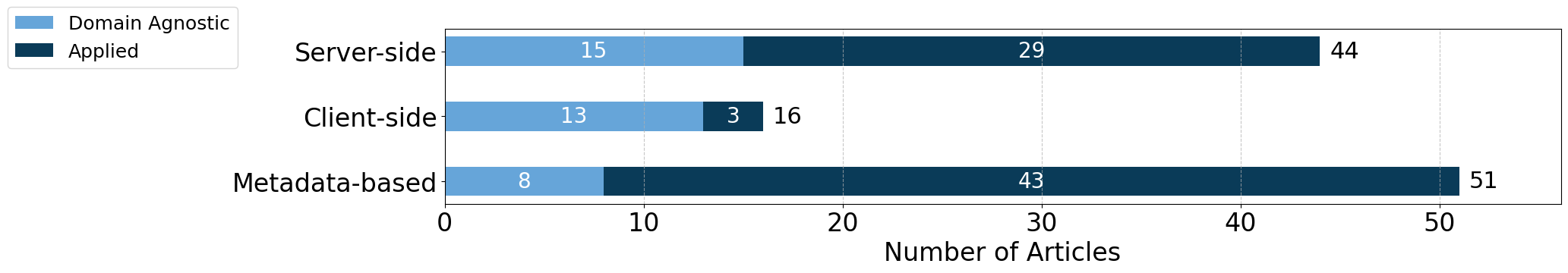}
    \caption{Comparison of clustering strategies between domain agnostic (articles reviewed in section~\ref{sec:clf_classification}) and applied domain-specific CFL studies (section~\ref{sec:clf_classification}).}
    \label{fig:applied_cfl}
\end{figure}

This section examines how CFL has been applied across specific domains, including IoT, mobility, energy, and healthcare. Unlike domain agnostic methods reviewed in section~\ref{sec:clf_classification}, which are predominantly server-side or client-side, applications overwhelmingly favor metadata-based clustering. As shown in Figure~\ref{fig:applied_cfl}, metadata-driven approaches are substantially more prevalent, highlighting their practicality in domain-specific settings. This preference for efficiency over strict privacy often results in solutions that are difficult to generalize.

Given the large number of domain-specific studies, a high-level summary and classification according to our taxonomy is provided in a supplement material~\ref{sec:appendix}. 

\subsubsection{Internet of Things and Networking}
\label{sec:IOT}
In IoT and networking environments, when device heterogeneity and resource constraints are critical, CFL aims to balance system and statistical heterogeneity. Many methods rely on structural metadata—such as resource efficiency, network topology, spectrum characteristics, or geographic information~\citep{sun2020adaptive,xu2023clustered,wasilewska2021federated,yao2023detection,lee2023energy,ali2025adaptive}—to guide clustering decisions. While some works use traditional server-side mechanisms with K-Means or DBSCAN on model weights~\citep{guo2024rec,fan2024taking,yu2025ddpg,chien2025density,he2023three,cao2025ap} or IFCA-like approaches~\citep{he2023clustered,sami2023over}, client-side solutions remain rare due to the high computational demand on edge devices. Although domain-specific tasks like network anomaly detection and reinforcement learning are explored~\cite{saez2023clustered,fan2024taking,wei2025toward,baek2025multifedrl}, many applied studies still use conventional image classification benchmarks like MNIST or CIFAR-10~\cite{he2023clustered,sami2023over,sun2020adaptive,lin2024rethinking,xu2023clustered,hamood2023clustered,he2023three,fotohi2024lightweight,hamood2024efficient,guo2024rec,zhao2023ensemble,yu2025ddpg,chien2025density}. This domain demonstrates CFL’s adaptability to resource-constrained scenarios, but highlights the tension between system efficiency and models performance.

\subsubsection{Mobility and Intelligent Transportation}
\label{sec:mobility}
The mobility domain, which includes vehicular networks and ITS, is characterized by dynamic connectivity and spatiotemporal variability~\cite{pei2024unveiling}. Metadata-based approaches dominate. They leverage shared information like geographic location, trajectory similarity, or mobility patterns to form clusters for tasks such as traffic forecasting and POI prediction~\cite{zhang2021communication,liu2020privacy,liao2024predicting,taik2022clustered}. Some other works also employ embeddings~\cite{gummadi2024fed,belhadi2025knowledge} as metadata. Server-side methods often rely on K-means or MTCFL~\cite{sattler2020clustered,ouyang2022clusterfl,zhang2021communication,wang2024contract,ye2023adaptive,ouyang2021clusterfl,lu2025clustered}. However, no client-side approaches were identified, even though vehicular and ITS clients may possess higher computational capabilities than standard IoT devices. 
Although domain-specific datasets (e.g., time-series traffic flow) and tasks (e.g., fault classification, HAR) are used~\cite{li2024echopfl,lu2024federated,belhadi2025knowledge}, a significant gap remains between the theoretical challenge of mobility and current CFL implementations. Although mobility inherently induces continuous statistical distribution drift (Data distribution change over time), none of the reviewed works employs dynamic mechanisms described in Section~\ref{sec:newcomers_dynamic}. Instead, adaptability in current literature is largely reactive to network topology (like in Taik et al.~\cite{taik2022clustered}) changes rather than statistical, highlighting a critical direction for future research in mobility-aware CFL.
 
\subsubsection{Energy}\label{sec:energy}

CFL is well-suited to the energy sector, where diverse user profiles naturally emerge. Clients typically correspond to different companies, households, or smart grids, each with its own specific consumption behaviors. CFL can exploit contextual similarities to create specific models.

 Most studies use metadata-based CFL with \textbf{time-series (TS)} forecasting. Geographical coordinates, daily load profiles, or statistical information are well suited as metadata to detect profiles  \cite{savi2021short,yoo2022fuzzy,he2023privacy,shi2023enabling,dogra2023consumers,liu2025cfl,ali2025explainable,wang2024domain}. For instance, Liue et al. \cite{liu2025cfl} employs privacy-preserving \textbf{Dynamic Time Warping (DTW)} for energy resource forecasting, while Ali et al. \cite{ali2025explainable} introduces an explainable CFL framework based on shared local dataset Shapley values for solar power prediction. Other uses CFL on load profiles or Markov transition matrices for short-term energy forecasting and load monitoring \cite{yoo2022fuzzy,dogra2023consumers,shi2023enabling}.

Server-side is less frequent, mostly applied to fault diagnosis and demand forecasting  \cite{zhou2025clustered,qin2023federated,cui2023federated,briggs2022federated}. Client-side remains rare, only emerging in a load monitoring framework (Wang et al.~\cite{wang2023blockchain}). Metadata-based once again overwhelmingly dominates, underscoring reliance on contextual information to optimize energy models—though often at the cost of reduced privacy.

\subsubsection{Healthcare}
\label{sec:healthcare}
Healthcare is among the most privacy-sensitive and heterogeneous domains, making CFL a promising paradigm for learning across distributed medical centers~\cite{qayyum2022collaborative, manthe2024federated, gao2025fedpc, niu2025fedcgp}. Interestingly, metadata-based approaches remain overrepresented, relying on sharing low-dimensional feature representations for clustering, such as latent representations from local encoders (e.g., ClusterGAN~\cite{jiang2023low}) or dataset-level embeddings~\cite{hsu2024cluster,cheema2025clustered,rafi2024tree}. A smaller number of studies explore server-side clustering strategies inspired by MTCFL~\cite{manthe2024federated, shaik2024clustered,guo2024molcfl,xie2021federated}. While medical image classification remains a common task~\cite{qayyum2022collaborative, jiang2023low, gao2025fedpc, cao2025ap}, CFL has been extended to complex tasks like molecular graph generation~\cite{guo2024molcfl}, 3D medical image segmentation~\cite{manthe2024federated}, and temporal physiological data analysis~\cite{yang2025clustering, cheema2025clustered, rafi2024tree, qiu2025fedkdc, shaik2024clustered}, demonstrating its flexibility across different healthcare data types.

\subsubsection{Summary and Outlook}
\label{sec:app_summary}
Beyond the major fields discussed above, CFL is also finding applications in advertising, blockchain, robotics, and speech processing~\cite{su2024empowering,yu2023federated,wu2024blockchain,wei2023edge,xiao2024knowledge,lee2024language,koppelmann22_interspeech,farahani2023toward}. The broad applicability of CFL across these diverse domains reinforces the real-world utility of the paradigm. However, a key observation is the methodological fragmentation: application-oriented studies overwhelmingly rely on metadata-based clustering strategies, while client-side approaches are almost nonexistent. This preference prioritizes communication efficiency and computational cost over privacy, which is particularly concerning in sensitive domains like healthcare. While these metadata-based approaches provide valuable, context-dependent insights, their reliance on domain-specific auxiliary data makes them difficult to generalize. Future research should encourage the adoption of client-side methods in settings where device computational capacity is sufficient, as they offer competitive performance with stronger privacy guarantees.

\section{Clustered X Federated Learning: Beyond Statistical Heterogeneity}
\label{sec:other_use_cases}

One key characteristic of Core Clustered Federated Learning (CFL) is its primary objective: mitigating non-IID data distributions by training multiple specialized models. In contrast, several recent works introduce the notion of \textbf{Clustered X Federated Learning} (borrowing the terminology from Liu et al.~\cite{fxl}), where X refers to architectural departures (e.g., decentralized, hierarchical, split, or resource-aware). These variants are fundamentally driven by architectural needs or system efficiency, not by the need to model distinct statistical distributions. We review these variants here to provide a clear disambiguation from Core CFL.

The following categories utilize clustering for operational purposes, contrasting them with the Core CFL objective of producing multiple specialized models.

\subsection{Clustered Decentralized FL} These adopt decentralized designs where clients communicate through direct interactions without the presence of a server. A the FL paradigm is changed completely, cluster formation is implicit and emergent, driven by iterative similarity among neighboring models rather than explicit server-side partitioning. Examples include GTV-CFL~\cite{GTV} and FedCBO~\cite{FedCBO}, which encourage local agreements; ACNM~\cite{li2022towards}, which adjusts communication links based on proximity; and blockchain-based coordination methods~\cite{xue2025clustered}. While they result in multiple specialized models, their decentralized architecture distinguishes them from the centralized Core CFL paradigm.

\subsection{Clustered Hierarchical FL} This category combines clustering with Hierarchical Federated Learning (HFL), where client results are first aggregated locally at an intermediate node before being sent to the global server. Although these approaches borrow CFL-related terminology, the role of clustering is fundamentally different from core CFL: clusters are created to optimize network performance rather than to separate client groups with distinct data distributions. Consequently, these methods universally maintain a \textbf{single global model} shared across the hierarchy, failing the multi-model requirement of Core CFL. Clustering in HFL is used to manage operational aspects within the multi-tier architecture. For example, Zhou et al.~\cite{zhou2023hierarchical} and Wang et al.~\cite{wang2024social} use clustering based on social-context information or privacy needs to facilitate participant selection or local aggregation, but the final output is a single global model. Similarly, Feng et al.~\cite{feng2022mobility} group clients based on mobility patterns to stabilize connectivity, while Wen et al.~\cite{wen2024dynamic} use dynamic clustering to adapt to rapidly changing network topology in Internet-of-Vehicles scenarios. The objective in all these cases is efficient aggregation and scheduling at the edge layer, with no intent to specialize models based on statistical heterogeneity. Overall, these Clustered Hierarchical FL methods do not leverage clustering to model statistical heterogeneity or to produce multiple specialized models.

\subsection{Clustered Split FL} This category combines CFL with Split Federated Learning (SFL). In SFL, the neural network is split into two parts: one part runs on each client device, and the other part runs on the server. During training, clients process their data through their portion of the model and send the resulting intermediate output to the server, which continues the forward and backward passes. This setup differs significantly from standard Federated Learning, where each client trains the entire model locally.

Because of this architectural split, these methods fall outside Core CFL. Their purpose is not to identify groups of clients needing different models, but rather to make SFL more efficient. Clustering here is mainly operational: grouping clients helps reduce communication delays, avoid slow devices (stragglers), or better organize how intermediate outputs are exchanged. In all cases, these approaches still aim to train one single global model, not several specialized ones.

Clustered SFL differs from Core CFL in two essential ways. First, the learning process depends on model partitioning rather than full-model aggregation, which changes communication patterns and optimization behavior. Second, clustering is used to manage practical limitations of SFL rather than to address statistical differences among clients. For example, Cheng et al.~\cite{cheng2023cheese} optimize network topology so that intermediate outputs can be transmitted efficiently, while Wazzeh et al.~\cite{wazzeh2024crsfl} cluster clients according to their hardware resources to prevent stragglers. Even Arafeh et al.~\cite{arafeh2025efficient}, although targeting non-IID IoT environments, still use clustering mainly to make the split architecture more efficient while training a single global model.

\subsection{Clustered Resource-Aware FL} This grouping of Clustered X variants focuses specifically on optimizing operational logistics. These methods are explicitly identified by terms such as Resource-Aware, Asynchronous, or Straggler-Aware FL, where the clustering criteria are driven by system-level constraints (e.g., resource availability, timing, or network topology) rather than statistical data distribution. The objective is to uniformly improve the scalability and stability of training a single global model. Several approaches focus on hardware and network limitations. For instance, Abdulrahman et al.~\cite{abdulrahman2024cras} introduce CRAS-FL , designed for vehicular networks, where clustering groups vehicles based on communication stability. Similarly, Mughal et al.~\cite{mughal2024adaptive} propose a framework to dynamically assign resource-constrained IoT devices to edge nodes based on resource availability. Other works address timing and scheduling challenges: Liu et al.~\cite{liu2024adaptive} introduce Straggler-Aware FL, using adaptive clustering to group clients by computation speed to mitigate delays. Qin et al.~\cite{qin2025collaborative} CAFL groups medical sensors based on communication quality to ensure reliable transmission times. Finally, Lu et al.~\cite{lu2023auction} apply clustering within an Auction-Based incentive mechanism to manage participation costs, while Wu et al.~\cite{wu2024fcer} employ Clustered Topology-Based edge selection to optimize data routing paths. Similarly, Hu et al.~\cite{hu2024sparsified} propose SRP-pFed, which utilizes clustering to optimize communication efficiency within Clustered personalized FL framework. This introduces a 'Sparsified Partial' update strategy where clients are clustered based on their optimal model 'update rate' (the ratio of parameters shared with the server). By grouping clients that require similar levels of global information, the system significantly reduces communication costs while maintaining \textbf{a single global model}. 

All the variants reviewed in the section, despite using the term "clustered FL", are fundamentally driven by architectural needs or system efficiency and are not primarily designed to solve the statistical heterogeneity problem through model specialization. By maintaining a \textbf{single global model} or by fundamentally altering the standard FL architecture, these Clustered X FL variants serve crucial roles in solving system heterogeneity but lie once again outside the scope of Core CFL for non-IID data specialization.

\section{Lessons Learned and Future Directions}
\label{sec:future_directions}

Our review reveals a significant gap between CFL theory and practice. While theoretical studies prioritize server-side and client-side clustering for data privacy, real-world applications in IoT, Mobility, and Energy overwhelmingly favor metadata-based approaches because practitioners prioritize efficiency and stability over strict privacy constraints. This separation means general algorithms are rarely tested under real-world constraints, and applications experiments frequently compare new algorithms only to older foundational methods like MTCFL~\cite{ghosh2019robust} or IFCA~\cite{IFCA}, rather than modern baselines. Furthermore, ambiguity persists in the terminology, where "Clustered X FL" is used for operational grouping, rather than to solve the core non-IID data problem.

Several key research directions emerge. First, evaluation must evolve toward a multi-criteria comparison framework aligned with the proposed taxonomy. Performance alone is insufficient, as each category imposes distinct trade-offs: server-side methods incur heavy centralized computation, client-side approaches exacerbate communication costs and the straggler effect, and metadata-based methods compromise privacy. Comparative studies must explicitly acknowledge and quantify these trade-offs rather than relying solely on performance metrics. Second, future evaluations must incorporate richer heterogeneity scenarios, with particular regard to Quantity Skew and Target Distribution Skew. Recent studies by Guo et al.~\cite{FedRC,HCFL+} highlight that existing CFL approaches struggle with label imbalance, leading to clusters that ignore underlying conceptual relationships. Including personalization methods in CFL may offer a more suitable alternative, and future research should investigate this possibility rigorously. Third, we emphasize the need to decouple the multi-tier CFL framework. Current methods are introduced as indivisible packages, simultaneously specifying initialization, clustering, dynamics, and personalization. Future work should conduct extensive ablation studies to isolate the impact of individual tiers, enabling both more interpretable research and the construction of modular CFL systems tailored to specific settings. Fourth, application-oriented CFL research, which currently centers heavily on metadata-based approaches, must establish a transparent performance–privacy trade-off analysis. Authors should explicitly quantify the privacy implications of each metadata component to justify its use. Conversely, client-side CFL approaches deserve more attention in application domains with substantial device computational capacity, as distributing the clustering burden can substantially improve scalability. Lastly, we encourage authors working on system-oriented variants (Hierarchical FL, Split FL, Resource-Aware FL) to adopt precise terminology rather than the ambiguous Clustered X FL label. Preserving terminological clarity is essential to maintain the historical definition of Core CFL as methods specifically designed to mitigate non-IID heterogeneity by producing multiple specialized models.

\section{Conclusion}\label{sec:conclusion}

This survey provides a unified conceptual foundation and a principled taxonomy for Clustered Federated Learning (CFL). We defined Core CFL as the set of methods explicitly designed to mitigate statistical heterogeneity by training multiple specialized models. The proposed taxonomy organizes existing methods into server-side, client-side, and metadata-based categories, detailing their respective trade-offs: server-side methods preserve strong privacy but lack scalability; client-side approaches distribute computation but burden devices; and metadata-based methods achieve efficiency but compromise privacy. We reviewed complementary CFL challenges, analyzed how heterogeneity taxonomies structure evaluation, and clarified the critical distinction between Core CFL and Clustered X FL variants. The latter, encompassing Hierarchical FL, Split FL, and Resource-Aware FL, employ clustering for architectural or system efficiency reasons while retaining a single global model objective, placing them outside the scope of Core CFL. We hope this unified framework supports the development of the next generation of rigorous and practically impactful CFL research.
\bibliographystyle{plainnat} 
\bibliography{refs} 
\newpage

\appendix
\section{Supplementary Material: High-level summary and classification of applicative CFL papers by domains}\label{sec:appendix}

In the following tables, we detail the clustering information and methodology used in each applied CFL paper. The 'CFL classification' column classifies each approach according to our taxonomy (Section~\ref {sec:cfltaxonomy}). For server-side and client-side approaches, the 'Clustering Strategy/Metadata' column briefly indicates the mechanism used and whether the method follows a known CFL baseline from Section~\ref{sec:clf_classification}. For metadata-based CFL, this column specifies the metadata shared with the server, such as geographical coordinates, resource availability or feature statistics. Finally, the 'Datasets \& Task' column lists the datasets, data types, and tasks evaluated.

\begin{table*}[ht!]
\tiny
\setlength{\tabcolsep}{5pt}
\renewcommand{\arraystretch}{0.9}
\centering
\caption{Applicative CFL papers in the IoT/Networking domain.}
\label{tab:cfl_iot}
\resizebox{\textwidth}{!}{
\begin{tabular}{lp{3cm}p{3cm}l}
\toprule
\textbf{CFL classification} & \textbf{Clustering Strategy/metadata} & \textbf{Datasets \& Task} & \textbf{Reference} \\
\midrule
Server-side & MTCFL-like \cite{sattler2020clustered} & Gotham IoT Testbed – Anomaly Detection (Time Series) & Saez et al. 2023 \cite{saez2023clustered} \\

Server-side & MTCFL-like \cite{sattler2020clustered} & CIFAR-10 – Image Classification & Hamood et al. 2023 \cite{hamood2023clustered} \\

Server-side & FlexCFL-like \cite{duan2021flexible} & FEMNIST – Image Classification & He et al. 2023 \cite{he2023three} \\

Server-side & Coalition Formation Game with model cosine similarity & MNIST, FEMNIST, CIFAR-10, California Housing – Classification/Regression & Zhao et al. 2023 \cite{zhao2023ensemble} \\

Server-side & Affinity Propagation + Cosine Similarity & Sent140, Fashion-MNIST, FEMNIST, CIFAR-10 – Text/Image Classification & Fotohi et al. 2024 \cite{fotohi2024lightweight} \\

Server-side & MTCFL-like \cite{sattler2020clustered}& FEMNIST, CIFAR-10 – Image Classification & Hamood et al. 2024 \cite{hamood2024efficient} \\

Server-side & K-means on weights & MNIST, CIFAR-10, EMNIST – Image Classification & Guo et al. 2024 \cite{guo2024rec} \\

Server-side & K-means on weights & IoT-23, IoTAnalytics – Anomaly Detection (Time Series) & Fan et al. 2024 \cite{fan2024taking} \\

Server-side & K-means on weights& MNIST, FMNIST, CIFAR-10 – Image Classification & Yu et al. 2025 \cite{yu2025ddpg} \\

Server-side & DBSCAN on weights & FEMNIST, CIFAR-10, Shakespeare – Image/Text Classification & Chien et al. 2025 \cite{chien2025density} \\

Client-side & IFCA-like \cite{IFCA} & MNIST, FMNIST, FEMNIST – Image Classification & He et al. 2023 \cite{he2023clustered} \\

Client-side & IFCA-like \cite{IFCA} & MNIST, CIFAR-10 – Image Classification & Sami et al. 2023 \cite{sami2023over} \\

Metadata-based & Computing power & MNIST – Image Classification & Sun et al. 2020 \cite{sun2020adaptive} \\

Metadata-based & Spectrum Sensing Metrics & Synthetic – Image Classification & Wasilewska et al. 2021\cite{wasilewska2021federated} \\

Metadata-based & Load/Geographical data & Simulation Data – Attack Classification & Yao et al. 2023 \cite{yao2023detection} \\

Metadata-based & Quality of Service / popularity & MovieLens – Recommendation & Huang et al. 2023 \cite{huang2023federated} \\

Metadata-based & Utility Balance & MNIST, CIFAR-10, CIFAR-100 – Image Classification & Xu et al. 2023 \cite{xu2023clustered} \\

Metadata-based & Edge-association & MNIST – Image Classification & Lee et al. 2023 \cite{lee2023energy} \\

Metadata-based & Temporal features & Gas Pipeline ICS, UNSW-NB15 – Intrusion Detection & Shan et al. 2023 \cite{shan2023cfl} \\

Metadata-based & Soft prediction vector & Fashion-MNIST, CIFAR-10, SVHN – Image Classification & Li et al. 2024 \cite{li2024personalized} \\

Metadata-based & Local model performance and labels balance & UNSW-NB15, CIC-IDS2017/18/19, BCCC-DDoS2024 – Anomaly Detection & Wei et al. 2025  \cite{wei2025toward} \\

Metadata-based & Route, movement type & Ray-tracing / MATLAB Comm. Toolbox – Channel frame prediction / beam tracking & Ali et al. 2025 \cite{ali2025adaptive} \\

Metadata-based & Environment/service type & IoT Simulation – Reinforcement Learning & Baek et al. 2025 \cite{baek2025multifedrl} \\

\bottomrule
\end{tabular}}
\end{table*}

\begin{table*}[ht!]
\tiny
\setlength{\tabcolsep}{5pt}
\renewcommand{\arraystretch}{0.9}
\centering
\caption{Applicative CFL papers in the Mobility domain.}
\label{tab:cfl_mobility}
\resizebox{\textwidth}{!}{
\begin{tabular}{lp{3cm}p{3cm}l}
\toprule
\textbf{CFL classification} & \textbf{Clustering Strategy/metadata} & \textbf{Datasets \& Task} & \textbf{Reference} \\
\midrule

Server-side & Kmeans-like on weights & HARBox; UWB; IMU; Depth gesture – HAR & Ouyang et al. 2021 \cite{ouyang2021clusterfl} \\

Server-side & FL+HC-like \cite{briggs2020federated} & PeMS, METR-LA – Traffic Forecasting & Zhang et al. 2021 \cite{zhang2021communication} \\

Server-side & MTCFL-like \cite{sattler2020clustered} & HAR datasets – HAR & Ouyang et al. 2022 \cite{ouyang2022clusterfl} \\

Server-side & Adaptive soft clustering using Gumbel–Softmax gating & GeoLife, Gowalla – Next POI Recommendation & Ye et al. 2023 \cite{ye2023adaptive} \\

Server-side & Kmeans on weights & MNIST, CIFAR-10, Belgium TSC – Classification & Wang et al. 2024 \cite{wang2024contract} \\

Server-side & FL+HC-like \cite{briggs2020federated}& Fashion-MNIST; CIFAR-10; SVHN – Classification & Lu et al. 2025 \cite{lu2025clustered} \\

Metadata-based & Geographic location & PeMS – Traffic Forecasting (Time Series) & Liu et al. 2020 \cite{liu2020privacy} \\

Metadata-based & Transmit power, link lifetime, channel state informations, velocity, dataset size and diversity (using gradients) & MNIST, FMNIST – Image Classification & Taik et al. 2022 \cite{taik2022clustered} \\

Metadata-based & Mobility factor& Simulation data – Reinforcement learning & Ye et al. 2024 \cite{ye2023federated} \\

Metadata-based & POI feature vectors & Chengdu – POI-based Forecasting & Liao et al. 2024 \cite{liao2024predicting} \\

Metadata-based & DBSCAN on mean embedding vectors & Outdoor robot navigation – Traversability Prediction & Gummadi et al. 2024 \cite{gummadi2024fed} \\

Metadata-based & Predictive-uncertainty scores & CWRU, Paderborn PU, ISU – Fault Diagnosis (Time Series) & Lu et al. 2024 \cite{lu2024federated} \\

Metadata-based & Data GMM & FMNIST, CIFAR-10, CelebA, BC, NER, ADE, SSM4H, Wisconsin, Yeast1, Vehicle, Shuttle, Car, KR/ Classification & Pei et al. 2024 \citep{pei2024unveiling} \\

Metadata-based & Local centroid & CIFAR-10; HAR-UCI; Ubisound/ Classification & Li et al. 2024 \cite{li2024echopfl} \\

Metadata-based & Contrastive embeddings & CityPersons, BDD100K, CCD – Pedestrian / anomaly detection & Belhadi et al. 2025 \cite{belhadi2025knowledge} \\

\bottomrule
\end{tabular}}
\end{table*}

\begin{table*}[ht!]
\tiny
\setlength{\tabcolsep}{5pt}
\renewcommand{\arraystretch}{0.9}
\centering
\caption{Applicative CFL papers in the Energy domain.}
\label{tab:cfl_energy}
\resizebox{\textwidth}{!}{
\begin{tabular}{lp{3cm}p{3cm}l}
\toprule
\textbf{CFL classification} & \textbf{Clustering Strategy/metadata} & \textbf{Datasets \& Task} & \textbf{Reference} \\
\midrule

Server-side & FL+HC-like \cite{briggs2020federated}& Low Carbon London – Forecasting (Time series) & Briggs et al. 2022 \cite{briggs2022federated} \\

Server-side & Kmeans on weights & Chinese gas companies – Forecasting (Time series) & Qin et al. 2023 \cite{qin2023federated} \\

Server-side & Kmeans on weights & Windfarm data – Fault diagnosis (Time series) & Zhou et al. 2025 \cite{zhou2025clustered} \\

Client-side & IFCA-like \cite{IFCA}& REFIT, REDD smart-meter – Forecasting (Time series) & Wang et al. 2023\cite{wang2023blockchain} \\

Metadata-based & Customers’ behavioral and socio-economic metadata & London Smart Meter Energy Data – Forecasting (Time series) & Savi et al. 2021 \cite{savi2021short} \\

Metadata-based & Geographical metadata & Solar + weather data – Forecasting (Time series) & Yoo et al. 2022 \cite{yoo2022fuzzy} \\

Metadata-based & Data centroid and avg load & Australian SGSC smart-meter – Forecasting (Time series)& He et al. 2023 \cite{he2023privacy} \\

Metadata-based & local SARIMA-parameters & China Carbon Accounting Database – Forecasting (Time series) & Cui et al. 2023 \cite{cui2023federated} \\

Metadata-based & Data-derived profiles & REFIT - Regression (Time series) & Shi et al. 2023 \cite{shi2023enabling} \\

Metadata-based & Signal/metadata extraction from TS & BC Hydro residential smart-meter – Forecasting (Time series) & Dogra et al. 2023 \cite{dogra2023consumers} \\

Metadata-based & Resource + data distribution statistics & FMNIST, CIFAR-10 – Image Classification & Yang et al. 2023 \cite{yang2023personalized} \\

Metadata-based & DTW-based centroids & BDG2 – Forecasting (Time series) & Liu et al. 2025 \cite{liu2025cfl} \\

Metadata-based & SHAP feature-contribution vectors & German Solar Farm – Forecasting (Time series) & Ali et al. 2025 \cite{ali2025explainable} \\

Metadata-based & Dataset Encoder output & Palo Alto EVCS; Boulder EVCS – Forecasting (Spatio-temporal Data) & Wang et al. 2024 \cite{wang2024domain} \\

\bottomrule
\end{tabular}}
\end{table*}
\begin{table*}[ht!]
\tiny
\setlength{\tabcolsep}{5pt}
\renewcommand{\arraystretch}{0.9}
\centering
\caption{Applicative CFL papers in Health.}
\label{tab:cfl_misc}
\resizebox{\textwidth}{!}{
\begin{tabular}{lp{3.2cm}p{3.3cm}l}
\toprule
\textbf{CFL classification} & \textbf{Clustering Strategy/metadata} & \textbf{Datasets \& Task} & \textbf{Reference} \\
\midrule

Server-side & Stoer–Wagner global min-cut (graph built with pairwise cosine similarities of client models) & MUTAG; BZR; COX2; DHFR; PTC\_MR; AIDS; NCI1; ENZYMES; DD; PROTEINS; COLLAB; IMDB-BINARY; IMDB-MULTI – Graph Classification & Xie et al. 2021 \cite{xie2021federated} \\

Server-side & MTCFL-like \cite{sattler2020clustered} & ESOL; QM9 – Generative molecular design (Graphs) & Guo et al. 2024 \cite{guo2024molcfl} \\

Server-side & K-means on weights & FeTS2022 (BraTS2021) – 3D brain tumor segmentation (3D image) & Manthe et al. 2024\cite{manthe2024federated} \\

Server-side & K-means/GMM on weights - HC+FL-like \cite{briggs2020federated} & PPG-DaLiA; UCI Drug Review – Medical HAR classification; Drug rating classification & Shaik et al. 2024 \cite{shaik2024clustered} \\

Server-side & MTCFL-like \cite{sattler2020clustered} & BraTS2018; Retinal vessel datasets (FIVES, STARE, DRIVE, IOSTAR, CHASE\_DB1, RAVIR) – Image segmentation  & Niu et al. \cite{niu2025fedcgp} \\

Server-side & FlexCFL-like \cite{duan2021flexible} with affinity propagation & MNIST; USPS; BloodMNIST; OrganAMNIST; OrganCMNIST; OrganSMNIST – Image classification & Cao et al. 2025\cite{cao2025ap} \\

Metadata-based & Based on modality & COVID-19 image data collection; Ultrasound COVID dataset (POCUS) – Image Classification & Quayyum et al. 2022 \cite{qayyum2022collaborative} \\

Metadata-based  & Bandwidth + dataset size + freshness of data & CASAS smart home sensor datasets – Classification (HAR) & Singh et al. 2022 \cite{singh2022federated} \\

Metadata-based & Features distribution & MIT Drivedb; AffectiveRoad – Binary driver-stress classification & Rafi et al. 2024 \cite{rafi2024tree} \\

Metadata-based & Clients token-frequency embeddings & ADReSS (DementiaBank) – Automatic Speech Recognition  (Audio) & Hsu et al. 2024 \cite{hsu2024cluster} \\

Metadata-based & Dataset extracted Embeddings  & WESAD wearable stress dataset – Classification (Time Series) & Jiang et al. 2023 \cite{jiang2023low} \\

Metadata-based & Daily carbohydrate (CHO) intake & UVA/Padova T1D Simulator – Forecasting (Time Series) & Yang et al. 2025 \cite{yang2025clustering} \\

Metadata-based & PCA vectors of clients features & Damage Simulation Data – Classification & Cheema et al. 2025 \cite{cheema2025clustered} \\

Metadata-based & Dataset prototype vectors & MedMNIST v2 (BloodMNIST, DermaMNIST) – Image Classification & Gao et al. 2025 \cite{gao2025fedpc} \\

Metadata-based & Clients Soft predictions & SEED; SEED-IV; SEED-FRA; SEED-GER – Emotion classification & Qiu et al. 2025 \cite{qiu2025fedkdc} \\

\bottomrule
\end{tabular}}
\end{table*}

\begin{table*}[ht!]
\tiny
\setlength{\tabcolsep}{5pt}
\renewcommand{\arraystretch}{0.9}
\centering
\caption{Applicative CFL papers in Advertising, Recommendation, Speech, and Robotics domains.}
\label{tab:cfl_other}
\resizebox{\textwidth}{!}{
\begin{tabular}{lp{3.2cm}p{3.3cm}l}
\toprule
\textbf{Approach} & \textbf{Clustering Metric / Strategy} & \textbf{Datasets \& Task} & \textbf{Reference} \\
\midrule

Server-side & MTCFL-like \cite{sattler2020clustered} & MNIST; CIFAR-10; iNaturalist-Geo; FEMNIST; Structured Non-IID FEMNIST; VisDA2017; iWildCam – Image classification & Wei et al. 2023 \cite{wei2023edge} \\

Server-side & K-means on weights & CPHPD (Persian dialect speech) – Speech recognition classification (Audio) & Farahni et al. 2023 \cite{farahani2023toward} \\

Server-side & Kmeans-like on weights & Ali Tianchi Display Ad Click; Kaggle CTR – Click rate prediction & Su et al. 2024 \cite{su2024empowering} \\

Server-side & DBSCAN on weights & LibriSpeech; UserLibri – Speech recognition & Lee et al. 2024 \cite{lee2024language} \\

Metadata-based & Clients Autoencoder outputs & Speech recognition & Koppelmann et al. 2022 \cite{koppelmann22_interspeech} \\

Metadata-based & Clients local click history (news) & Adressa; MIND – News click prediction / recommendation ranking & Yu et al. 2023 \cite{yu2023federated} \\

Metadata-based & Clients dataset statistics + training-time & CIFAR-10; CIFAR-100; Shakespeare; Cora; CiteSeer – Image classification; character prediction; graph node classification & Wu et al. 2024 \cite{wu2024blockchain} \\

Metadata-based & Validation accuracy score vectors & Proprietary robotic-assembly dataset – Fault diagnosis & Xiao et al. 2024 \cite{xiao2024knowledge} \\

\bottomrule
\end{tabular}}
\end{table*}

\normalsize

\end{document}